\documentclass[journal]{IEEEtran}
\usepackage{cite}
\usepackage{hyperref}

\ifCLASSINFOpdf
   \usepackage[pdftex]{graphicx}
\else
   \usepackage[dvips]{graphicx}
\fi
\usepackage{makecell}
\usepackage{booktabs}
\usepackage{xcolor}
\usepackage{multirow}

\usepackage{amsmath,amssymb}

\usepackage{pifont}

\usepackage{url}
\usepackage{dsfont}

\newcommand{\boldstart}[1]{\noindent\textbf{#1}}
\newcommand{\ie}{\textit{i}.\textit{e}., }
\newcommand{\eg}{\textit{e}.\textit{g}. }

\begin{document}
%
\title{Snipper: A Spatiotemporal Transformer for Simultaneous Multi-Person 3D Pose Estimation Tracking and Forecasting on a Video Snippet}
%
%
%

\author{Shihao~Zou,
        Yuanlu~Xu,
        Chao~Li,
        Lingni~Ma, 
        Li~Cheng,
        and~Minh~Vo
\thanks{Part of this work was done during Shihao Zou's internship at Meta Reality Labs. The results on JTA dataset were added after his internship. Minh Vo was at Meta Reality Labs during the development of this work.}
\thanks{S. Zou and L. Cheng are with the Department
of Electrical and Computer Engineering, University of Alberta, Canada. (E-mail: szou2@ualberta.ca, lcheng5@ualberta.ca).}
\thanks{Y. Xu, C. Li, and L. Ma are with the Meta Reality Labs, United States. (E-mail: yuanluxu@meta.com, chao.li@meta.com, lingni.ma@meta.com).}
\thanks{M. Vo is with Spree3D, United States. (E-mail: minh.vo@spree3d.com).}
\thanks{Manuscript received 08 Oct 2022; accepted 03 Feb 2023. Li Cheng is the corresponding author for this paper.}}

%
%

\markboth{Journal of \LaTeX\ Class Files,~Vol.~14, No.~8, August~2015}%
{Shell \MakeLowercase{\textit{et al.}}: Bare Demo of IEEEtran.cls for IEEE Journals}
%



\maketitle

\begin{abstract}
Multi-person pose understanding from RGB videos involves three complex tasks: pose estimation, tracking and motion forecasting. Intuitively, accurate multi-person pose estimation facilitates robust tracking, and robust tracking builds crucial history for correct motion forecasting. Most existing works either focus on a single task or employ multi-stage approaches to solving multiple tasks separately, which tends to make sub-optimal decision at each stage and also fail to exploit correlations among the three tasks. In this paper, we propose Snipper, a unified framework to perform multi-person 3D pose estimation, tracking, and motion forecasting simultaneously in a single stage. We propose an efficient yet powerful deformable attention mechanism to aggregate spatiotemporal information from the video snippet. Building upon this deformable attention, a video transformer is learned to encode the spatiotemporal features from the multi-frame snippet and to decode informative pose features for multi-person pose queries. Finally, these pose queries are regressed to predict multi-person pose trajectories and future motions in a single shot. In the experiments, we show the effectiveness of Snipper on three challenging public datasets where our generic model rivals specialized state-of-art baselines for pose estimation, tracking, and forecasting. Code is available at \href{https://github.com/JimmyZou/Snipper}{https://github.com/JimmyZou/Snipper}.  
\end{abstract}


%
\IEEEpeerreviewmaketitle

\section{Introduction}
\label{sec:intro}
Multi-person 3D pose understanding from RGB videos is a fundamental topic in computer vision, which mainly involves three complex tasks, namely multi-person pose estimation, tracking, and motion forecasting. 
These three tasks are all desired in a wide range of applications such as human action recognition and behavior analysis, pedestrian tracking, re-identification, human-computer interaction, and video surveillance~\cite{song2020richly,sun2021survey,munea2020progress,benedek2016lidar}. As an example of human behavior analysis in a crowded scene, multi-person pose estimation and tracking usually provides critical information for the accurate analysis, and motion forecasting further helps behavior forecasting and directs region-of-interest in the future for the system.

\begin{figure}[ptb]
    \centering
    \includegraphics[width=\columnwidth]{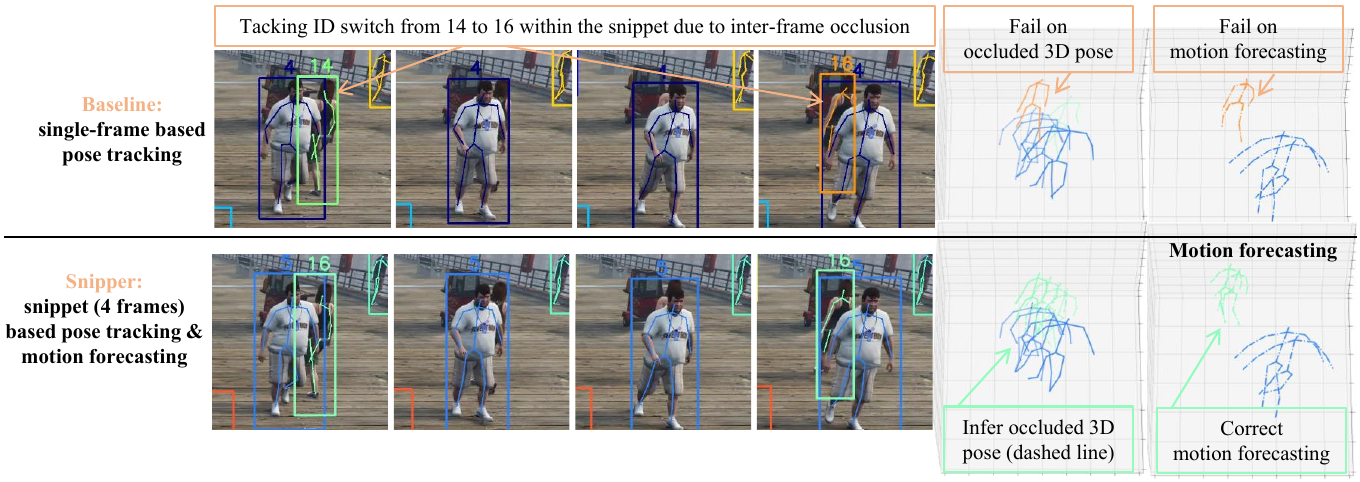}
    \caption{A practical yet challenging example to track human pose and motion forecasting. The baseline method typically adopts a tracking-by-detection schema based on single-frame inference, which fails under heavy occlusions. The proposed method Snipper resolves the ambiguity by processing video snippets with a single-stage framework instead. With the proposed attention module, Snipper aggregates spatiotemporal features shared by the three tasks. On a test set from JTA dataset with 11K 4-frame snippets that present inter-frame occlusion, Snipper obtains 84.2 3D-PCK and 62.3 MOTA (4-frame snippet) versus 80.1 3D-PCK and 59.5 MOTA (single-frame).
    }
    \label{fig:motivation}
\end{figure}

Earlier works either focus on a single task~\cite{wang2022ktn,li2022multiperson,gu2019multi,martinez2020efficient,sun2021survey,wei2019view,openpose,benzine2020pandanet,tu2020voxelpose,zou2022human}, or employ multi-stage approaches to solving multiple tasks separately~\cite{reddy2021tessetrack,rajasegaran2021tracking,xiao2018simple,fabbri2018learning,cao2020long}. However, for the challenging scenarios with heavy occlusions, such as the example in Fig.~\ref{fig:motivation}, the existing methods tend to fail in correct pose estimation and robust tracking of the occluded persons in a video, and further lead to inaccurate motion forecasting due to the misleading history information. This failure can be mainly attributed to three factors: i) temporal information is not considered in the single-task methods, especially for single-frame based multi-person pose estimation~\cite{wang2022ktn,li2022multiperson,martinez2020efficient,gu2019multi,openpose}; ii) multi-stage methods tend to make sub-optimal decisions when addressing the three tasks step by step, because the reasoning is usually not conducted in a joint space, such as earlier works~\cite{villegas2017learning,walker2017pose,gupta2018social,cao2020long} generally treat pose tracking as separate modules for the motion forecasting; and iii) the correlations among the three tasks are not fully exploited. Intuitively, accurately estimated multi-person poses facilitate robust tracking. Conversely, robust tracking provides informative region-of-interest context for the multi-person pose estimation, and it also builds the crucial history that enables sensible prediction of motion in the future.

To address the above limitations, we propose Snipper, a unified framework to simultaneously estimate, track, and forecast multi-person 3D poses on a video snippet of consecutive RGB frames. Snipper reasons the three tasks in a joint spcae and regresses 2.5D multi-person pose trajectories and their future motions from a snippet in a single stage. This framework is inspired by the query-based DETR framework~\cite{carion2020end,wang2021end,zhu2020deformable} for object detection, but with contributions in an efficient yet powerful spatiotemporal deformable attention module for fine-grained video understanding tasks. 

Specifically, our proposed attention mechanism adopts the sparse spatial deformable attention~\cite{zhu2020deformable} to process high-resolution multi-scale image features for better image-aligned tasks. 
However, na\"ive extension to video by regressing a space-time offset and directly sampling in 3D space is problematic as the interpolation in time domain is ill-defined without temporal correspondences. In addition, the image features at the same spatial position across frames often change due to object or camera motions.
To tackle this problem, we propose to restrict the temporal offset to pre-defined integer frame index and only regress spatial offset on those frames to aggregate per-frame image features in space and time. 
Unlike the self-attention in Carion et al.~\cite{carion2020end}, our proposed mechanism is efficient and also maintains the spatiotemporal relationship of multi-frame and multi-scale features when aggregating spatiotemporal features by deformable attention. 
Compared with spatial attention~\cite{zhu2020deformable}, our strategy further considers temporal features. This efficient yet powerful solution to the aggregation of multi-frame features is critical to resolve the information missing due to occlusions or motion blur within the snippet. We provide detailed discussion of these paradigms in Sec.~\ref{sec:3Dattention} and comparison in Sec.~\ref{sec:ablation} and Fig.~\ref{fig:compare_attention}. 

Based on the proposed spatiotemporal deformable attention module, we devise a deformable transformer for simultaneous completion of the three tasks. Concretely, given the multi-frame feature volume extracted by a CNN backbone, the encoder (Sec.~\ref{sec:encoder}) aggregates spatiotemporal feature via the attention module to update features of each voxel. Then the feature volume from the encoder is fed to the transformer decoder (Sec.~\ref{sec:decoder}) as the memory, from which multi-person pose queries can accumulate pose trajectory features via the same attention module.
Finally, these queries are regressed to predict multi-person pose trajectories in the observed frames and also motions for the future frames. (Sec.~\ref{sec:losses}) To achieve tracking in the entire video, we run Snipper on the overlapping snippets in a video, and associate the pose trajectories based on the common frame of two consecutive snippets. No further appearance descriptor is needed. 

\boldstart{Our contributions} are summarized as follows:
\begin{itemize}
    \item We propose Snipper, a unified framework for simultaneous multi-person 3D pose estimation, tracking, and motion forecasting from a video snippet. To our knowledge, the proposed framework is the first one that jointly solves these three tasks in a single stage.
    
    \item We propose an efficient yet powerful spatiotemporal deformable attention mechanism in the transformer to aggregate spatiotemporal information from multi-scale and multi-frame feature volumes. Its effectiveness and efficiency is discussed in Sec.~\ref{sec:3Dattention} and validated in Sec.~\ref{sec:ablation}, and our proposed spatiotemporal attention module is also general to other image-aligned video understanding tasks.
    
    \item We validate the proposed framework on three challenging datasets: JTA~\cite{fabbri2018learning}, CMU-Panoptic Studio~\cite{joo2017panoptic}, and PoseTrack18~\cite{PoseTrack}. We show that a generic Snipper model presents competitive performance on all three tasks of pose estimation, tracking, and motion forecasting compared with specialized baselines that tackle only one or two tasks.
\end{itemize}

\section{Related Works}
\boldstart{Multi-person pose estimation} from monocular images has been extensively investigated in the past few years. Existing methods can be divided into three categories: bottom-up~\cite{pishchulin2016deepcut,openpose,fabbri2020compressed,fabbri2018learning,zanfir2018deep,martinez2020efficient,li2022multiperson,gu2019multi}, top-down~\cite{chen2018cascaded,girdhar2018detect,moon2019camera,cheng2021monocular,tu2020voxelpose,wang2020hmor,reddy2021tessetrack,cheng2021graph,wang2022ktn} and single-stage~\cite{benzine2020pandanet,mehta2020xnect,nie2019single,sun2021monocular,wang2021mvp,jin2022single,wang2022distribution,shi2022end}.

\textit{Bottom-up} methods detect 2D joints first and estimate 2D or 3D pose with different association approaches such as integer linear program~\cite{pishchulin2016deepcut,zanfir2018deep}, or Part Affinity Fields~\cite{openpose}. In addition, Gu et al.~\cite{gu2019multi} formulates 3D pose estimation as a Perspective-N-Point optimization problem based on detected multi-person 2D poses via~\cite{openpose} and shows good performance with high efficiency for 3D pose estimation. Recently, Fabbri et al.~\cite{fabbri2020compressed} extends 2D heatmaps to 3D compressed volume for direct 3D joints detection for multi-person 3D pose estimation. There is also effort~\cite{martinez2020efficient} using depth map for efficient multi-person 2D pose estimation with CNN and knowledge distillation at multiple architecture levels. A most recent work~\cite{li2022multiperson} trains a Hourglass model~\cite{newell2016stacked} to predict multi-person 2D key-points heatmap with peak regularization and employs greedy key-point association to obtain multi-person 2D poses. Compared with the bottom-up methods, our unified framework does not depend on external 2D pose detector or require extra joints association step. Besides, our work is not limited to multi-person pose estimation, but achieves three tasks in a single stage with an end-to-end model.

\textit{Top-down} methods first detects the person bounding box and applies single person pose estimation on the cropped region. A pose proposal generator is employed in~\cite{rogez2019lcr} followed by a pose refinement regressor. RootNet~\cite{moon2019camera} infers multi-person 3D pose by detecting absolute 3D root localization first and then estimating root-relative single-person 3D pose. Wei et al.~\cite{wei2019view} presents a view-invariant hierarchical correction network on top of an initial estimated single-person 3D pose to learn the 3D pose refinement under consistent views, and then uses a view-invariant discriminative network to enforce high-level constraints over body configurations.
HMOR~\cite{wang2020hmor} encodes interaction information of multiple persons as the ordinal relations of depths and angles hierarchically. The recent effort~\cite{cheng2021graph} applies graph and temporal convolutional neural networks for multi-person pose estimation in a video. In general, top-down methods estimate more accurate poses than bottom up counterpart as the expense of more compute. Integrating bottom-up and top-down is considered in~\cite{cheng2021monocular} to complement each other. Multi-view top-down approaches are investigated in~\cite{tu2020voxelpose,reddy2021tessetrack} where human are detected and integrated from multi-view sources in a 3D volume and regressed to estimate 3D pose. A most recent effort~\cite{wang2022ktn} proposes knowledge transfer network to learn the 2D-3D correspondences for multi-person 3D dense pose estimation because of insufficient and imbalanced 3D labels. Our method differs from the top-down methods in that we do not need extra person detector to localize persons.

\textit{Single-stage} methods are emerging in recent years for both pose estimation~\cite{mehta2020xnect,nie2019single,zhen2020smap,benzine2020pandanet,jin2022single,wang2022distribution,shi2022end,benzine2021single} and parametric body mesh estimation~\cite{sun2021monocular}. Our framework also adopts this paradigm for its simplicity, while also considering multi-person tracking and motion forecasting in our pipeline.

\boldstart{Multi-person pose tracking} aims to track multi-person poses in a video. Recently, \cite{sun2021survey} provides a survey of multiple pedestrian tracking based on the tracking-by-detection framework. For 2D pose tracking, Girdhar et al.~\cite{girdhar2018detect} uses top-down approaches to estimate frame-based multi-person poses and then links predictions over time using bipartite matching. \cite{xiao2018simple,hwang2019pose} employ a similar top-down detection strategy but rely on flow-based similarity to do pose tracking. Wang et al.~\cite{wang2020combining} extends HRNet~\cite{sun2019deep} with temporal convolutions and shows impressive joint pose estimation and tracking. In contrast, Raaj et al.~\cite{raaj2019efficient} propose and efficient the bottom-up approach by extending the spatial affinity fields to spatiotemporal affinity fields in an RNN model. For 3D pose tracking, two multi-stage approaches are also proposed in~\cite{mehta2020xnect,zanfir2018monocular} where per-frame multi-person 3D pose estimation is followed by a temporal constraint optimization or fitting step. In contrast, ~\cite{fabbri2018learning,reddy2021tessetrack} aggregate temporal information within the model to estimate the multi-person pose trajectory. 
The most recent top-down works~\cite{rajasegaran2021tracking,rajasegaran2022tracking} achieve tracking by using 3D representation of people or predicting the future state of the tracklet, including 3D location, appearance, and pose. Besides, there are also efforts~\cite{zhang2022voxeltrack} exploring pose tracking with cross-view correspondence for occlusion-aware 3D tracking. Different from these multi-stage methods, our unified framework achieves 3D pose tracking and motion forecasting within a snippet in a single shot.

\begin{figure*}[ptb]
    \centering
    \includegraphics[width=\textwidth]{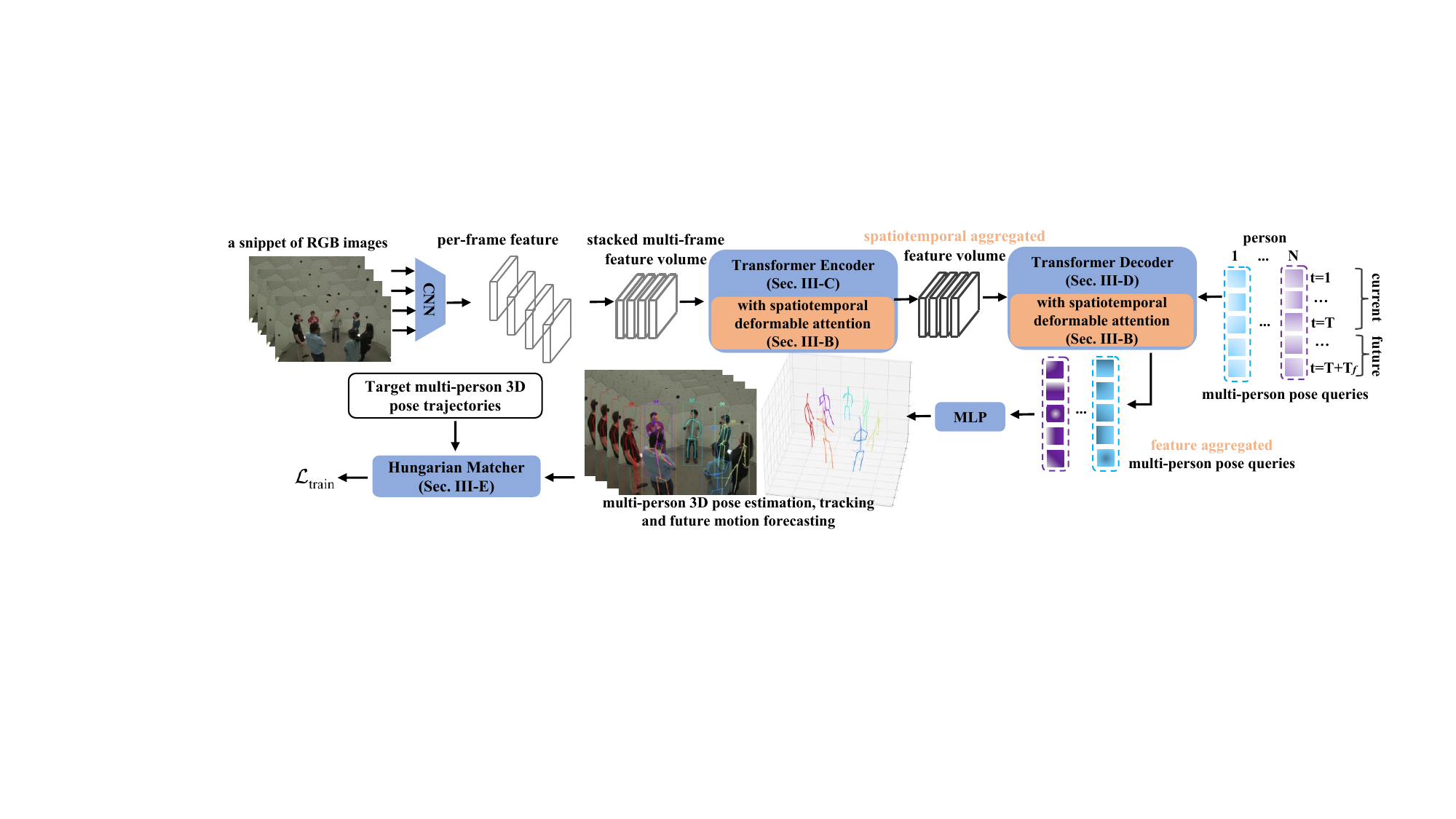}
    \caption{\textbf{Snipper overview.} Taking a snippet of $T$ consecutive RGB images as the input, CNN is used to extract per-frame image features, which are stacked into a multi-frame feature volume. This feature volume is fed into the transformer encoder (Sec.~\ref{sec:encoder}) that employs a novel spatiotemporal deformable attention (Sec.~\ref{sec:3Dattention}) to aggregate features.
    Given the spatiotemporally aggregated feature volume from the encoder, the transformer decoder (Sec.~\ref{sec:decoder}) aggregates pose features from it via spatiotemporal deformable attention. Those aggregated pose features are used to update pose queries for $N$ people of $T$ observed frames and $T_f$ future frames. 
    The updated queries are regressed to estimate each person's 3D pose tracking over $T + T_f$ frames. 
    We use Hungarian match to find an optimal permutation of a fix set of predicted 3D pose trajectories to a set of target trajectories to compute the multi-person pose losses for training. (Sec.~\ref{sec:losses}).
    }
    \label{fig:pipeline}
\end{figure*}

\boldstart{Transformer} is originally proposed in~\cite{vaswani2017attention} where self-attention is used and achieves the state-of-art results on many sequence-based tasks. DETR~\cite{carion2020end} and VisTR~\cite{wang2021end} are recent inspiring attempts to apply transformer in the end-to-end object detection and instance segmentation. 
Besides, multi-object tracking with Transformer has also been investigated in~\cite{zeng2021motr,meinhardt2022trackformer}.
However, due to the high compute requirement of the dot-product attention, both DETR and VisTR can only process low-resolution feature map which limits their accuracy. Deformable attention mechanism is proposed in~\cite{zhu2020deformable} to tackle this issue, showing strong accuracy for small objects detection. Transformer is recently applied to single person pose estimation, where self-attention module is directly applied to the joint or mesh vertex positions~\cite{lin2021end,zheng2021poseformer,li2022mhformer}, and also multi-frame image features~\cite{wan2021encoder,wang2021mvp}. The most recent work~\cite{shi2022end} has also employed transformer for multi-person pose estimation, where the query-based self-attention transformer is used to regress multi-person poses for a single frame. To the best of our knowledge, our method is the first to use the deformable attention mechanism for simultaneous multi-person pose estimation, tracking, and forecasting with a single-stage paradigm.

\section{Method}
Snipper is a unified framework that simultaneously addresses three tasks from an RGB video snippet of $T$ observed frames: multi-person 3D pose estimation, tracking, and forecasting for the future $T_f$ frames. It consists of four components: image features extractor, transformer encoder (Sec.~\ref{sec:encoder}), decoder (Sec.~\ref{sec:decoder}) and pose trajectory matching for training (Sec.~\ref{sec:losses}). A novel spatiotemporal deformable attention (\ref{sec:3Dattention}) is employed in both the transformer encoder and decoder to aggregate informative pose features shared by the three tasks. To achieve tracking in the entire video, we associate multi-person pose trajectories based on the common frame of two consecutive overlapping snippets with more details presented in the supplementary material. An overview of our pipeline is shown in Fig.~\ref{fig:pipeline}.

\subsection{Preliminary}
\boldstart{Star pose representation.} 
We represent the 3D human pose as $\mathbf{P}=\{\mathbf{J},\mathbf{V}, o\}$ where $\mathbf{J}=\{J_i: J_i \in \mathrm{R}^3\}_{i=1}^{N_J}$ is the set of joint locations, $\mathbf{V}=\{V_i: V_i \in [0, 1]\}_{i=1}^{N_J}$ is the joint visibility, and $o \in [0, 1]$ is the person occurrence probability. Each individual joint position ${J_i}$ is modeled by the offset ${J}_i^{\text{offset}} = \{\Delta x, \Delta y, \Delta d\}$ from the global root $J^{\text{root}} = \{x,y,d\}$, \ie $\ J_i = {J}^{\text{root}}+{J}_i^{\text{offset}}$, where $(x,y)$ are the 2D image location of the joint and $d$ is its depth to the camera center, respectively. 

\boldstart{Depth and joint offset normalization.} 
As the absolute root depth $d$ depends on the camera focal length $f_c$, we normalize the root depth by $\tilde d = d / f_c$, similar to~\cite{cheng2021graph}. In addition, the magnitude of 2D joint offset $(\Delta x, \Delta y)$ is in pixel distance and thus depends on the depth of the person. That is, a person's joint offset will become smaller if it moves far away from the camera, which tends to make the training in-stable. We mitigate this issue by normalizing the joint offset with the inverse of the normalized root depth, \ie $\Delta \tilde x = \Delta x \cdot \tilde d$ and $\Delta \tilde y = \Delta y \cdot \tilde d$. More details are described in the supplementary material. Thus, the magnitude of 2D normalized joint offset only depends on the pose of the person, which is more consistent across identities. During inference, we assume camera intrinsic is known with a fixed aspect ratio. Otherwise, we use a default focal length and pad the image to the predefined aspect ratio. Finally, a 3D joint position can be converted from the 2.5D joint representation. 

\subsection{Frame-Level Feature Extraction}
\label{sec:cnn}
Given an RGB video snippet of $T$ frames, CNN is used to extract per-frame features of size $H \times W \times C$. We stack these $T$ frame-level features through time and obtain the multi-frame feature volume $\mathbf{F}\in \mathbb{R}^{T\times H\times W\times C}$. Note that the multi-scale pyramid features $\{\mathbf{F}^l\}$ extracted by CNN can be easily applied in the followed Transformer Encoder and Decoder for fine-grained spatiotemporal feature extraction. Details are illustrated in Sec.~\ref{sec:3Dattention} and Fig.~\ref{fig:multiscale-attention}.

\subsection{Spatiotemporal Deformable Attention}
\label{sec:3Dattention}
Spatiotemporal deformable attention module is shown to produce more informative feature for pose tracking from the stack of multi-frame feature volume $\mathbf{F}\in\mathbb{R}^{T\times H \times W \times C}$. (validated in  Sec.~\ref{sec:ablation}) Such aggregation is crucial to mitigate the common inter-frame information missing such as self and partial occlusion. 

\begin{figure}[ptb]
    \centering
    \includegraphics[width=\columnwidth]{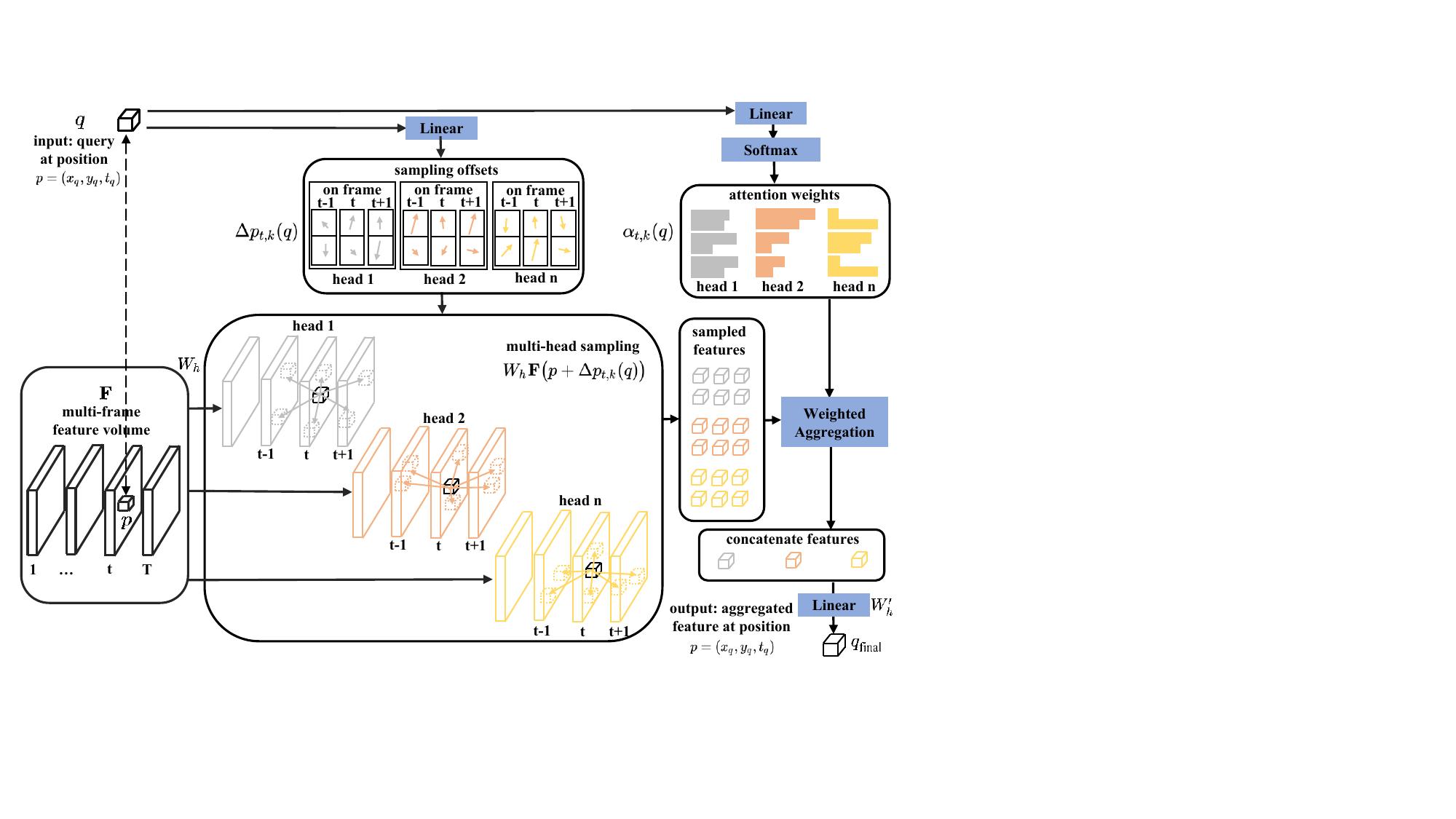}
    \caption{\textbf{Spatiotemporal deformable attention}. Given a volume of stacked multi-frame features, a learnable query $q$, specified at the position $(x_q, y_q, t_q)$, regresses multiple 2D offsets $\Delta p_{t, k}$ to sample features on a set of its neighboring frames $\mathbf{S}(t_q)$ and the attention weights $\alpha_{t, k}$ to aggregate features. 
    The feature volume is mapped to multiple attention heads with distinct linear layers in parallel, and the accumulated features are concatenated to form the final spatiotemporally aggregated features.}
    \label{fig:deformable-attention}
\end{figure}

We summarize our proposed spatiotemporal deformable attention in Fig.~\ref{fig:deformable-attention}. Let $q\in\mathbb{R}^C$ be the query specified at position $p=(x_q, y_q, t_q)$, where $x_q,\ y_q\in[0,1]$ is the normalized pixel spatial position and $t_q$ is the integer frame index of the query $q$. Then, $q$ is passed through two MLPs to regress 2D offsets $\Delta p_{t, k}(q)$ and corresponding attention weights $\alpha_{t, k}(q)$ normalized by the soft-max function. $t$ is an integer specifying the temporal frame in a pre-defined set of neighboring frames $\scriptsize \mathbf{S}(t_q)=\{t_q-1, t_q, t_q+1\}$, and $k$ indexes the offsets on each temporal frame. Note that we do not regress time offset but only 2D spatial offsets $\Delta p_{t, k}$ on each frame of $\mathbf{S}(t_q)$. We execute this process in parallel by multiple independent heads $h$ and form the final aggregated feature $q_{\text{final}}$ by passing the concatenated feature from each head through a linear layer,
\begin{equation}\small\begin{aligned}
    \label{eq:3dattention}
    q_{\text{final}} &= \sum_{h} W'_{h} \Big[ \sum\limits_{t,k} \alpha_{t,k}(q) \cdot W_{h} \mathbf{F}\big(p+\Delta p_{t,k}(q)\big) \Big], 
\end{aligned}\end{equation}
where $W_h$ and $W'_h$ are the parameters of linear layers.

\begin{figure}[ptb]
    \centering
    \includegraphics[width=\columnwidth]{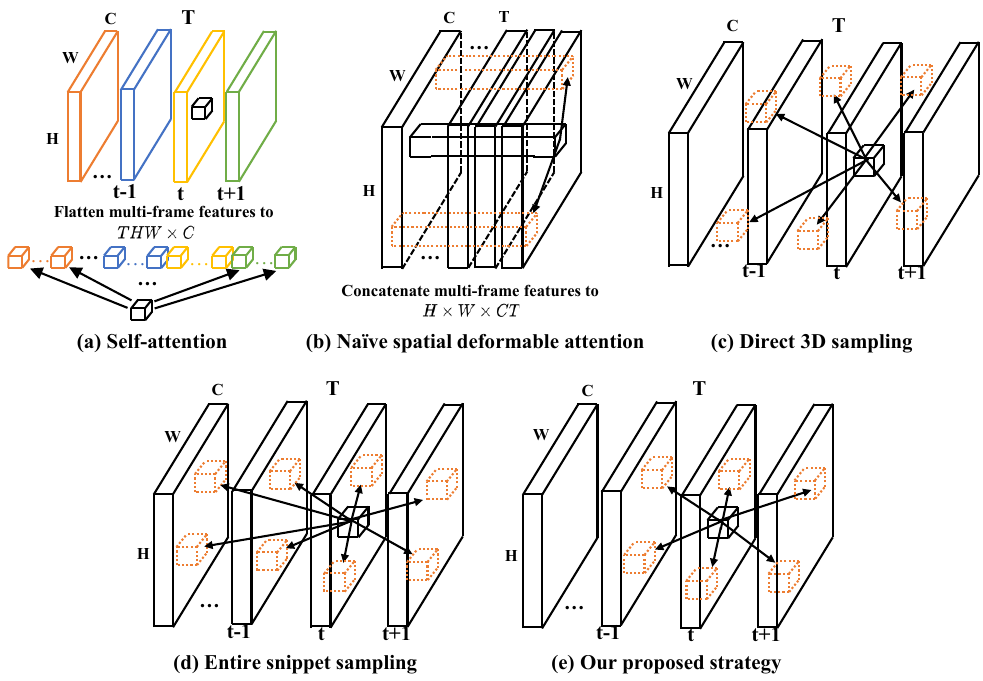}
    \caption{\textbf{Discussion of different attention mechanisms.} (a) Self-attention flattens multi-frame features to $\scriptsize THW\times C$ with attention applied on all $\scriptsize THW$ voxels. (b) Na\"ive spatial deformable attention concatenates temporal features to $\scriptsize CT\times H\times W$ with deformable attention applied on spatial space $\scriptsize H\times W$. (c) Direct 3D sampling regresses space-time offsets and directly samples in 3D space $\scriptsize T\times H\times W$, where the interpolation is performed in both spatial and temporal domain. (d) Entire snippet sampling performs attention over all the frames within the input snippet. (e) Our proposed spatiotemporal deformable attention samples on neighboring frames efficiently without performance reduction.}
    \label{fig:attention-discussion}
\end{figure}

\boldstart{Discussion.} There are several alternatives to implement spatiotemporal attention with details displayed in Fig.~\ref{fig:attention-discussion}: 
\begin{itemize}
    \item (a) \emph{Self-attention}. Follow VisTR~\cite{wang2021end}, we flatten multi-frame features to a 2D matrix of shape $THW\times C$ and applies attention to all $THW$ voxels. This attention mechanism is costly for high-resolution feature maps and also breaks the local spatiotemporal relationship for better image-aligned tasks.
    
    \item  (b) \emph{Na\"ive spatial deformable attention}~\cite{zhu2020deformable}. We reshape the feature volume $\mathbf{F}$ to the shape $H \times W \times CT$, where the channel size becomes $CT$ after concatenating multi-frame temporal features at the same image position. Then the spatial deformable attention is applied on the spatial domain $H\times W$ to aggregate spatiotemporal features. However, this na\"ive extension fails to consider object or camera motions within a video snippet. With these motions, image features at the same spatial position across frames often change, but the temporal features are still aggregated on the fixed spatial positions.
    
    \item (c) \emph{Direct 3D sampling}: This approach regresses space-time offsets and directly samples in the 3D space $T\times H\times W$, where the interpolation is performed in both spatial and temporal domain, \ie, $t\in[1, T]$ is fractional instead of integer frame index. However, the temporal interpolation is costly and ill-defined without known correspondences between frames such as optical flow, which leads to defects in the aggregated temporal features. 
    
    \item (d) \emph{Entire snippet sampling}: Another scheme is to sample on all frames of  the input snippet for the query at $(x_q, y_q, t_q)$, with offsets restricted on each temporal frame in the snippet.
    
    \item (e) \emph{Neighboring frame sampling (ours)}: Our approach limits the sampling range to be the immediate neighboring frames $\mathbf{S}(t_q)$. Despite the short temporal connection at each spatiotemporal deformable attention module, the temporal information is still fully accumulated due to the multiple layers of the attention module in the transformer encoder. Compared with the approaches mentioned above, our proposed mechanism requires less computation, but without any performance reduction. More results and analysis are presented in Sec.~\ref{sec:ablation} and Fig.~\ref{fig:compare_attention} that validates the effectiveness and efficiency of our proposed strategy. 
\end{itemize}

\begin{figure}[ptb]
    \centering
    \includegraphics[width=\columnwidth]{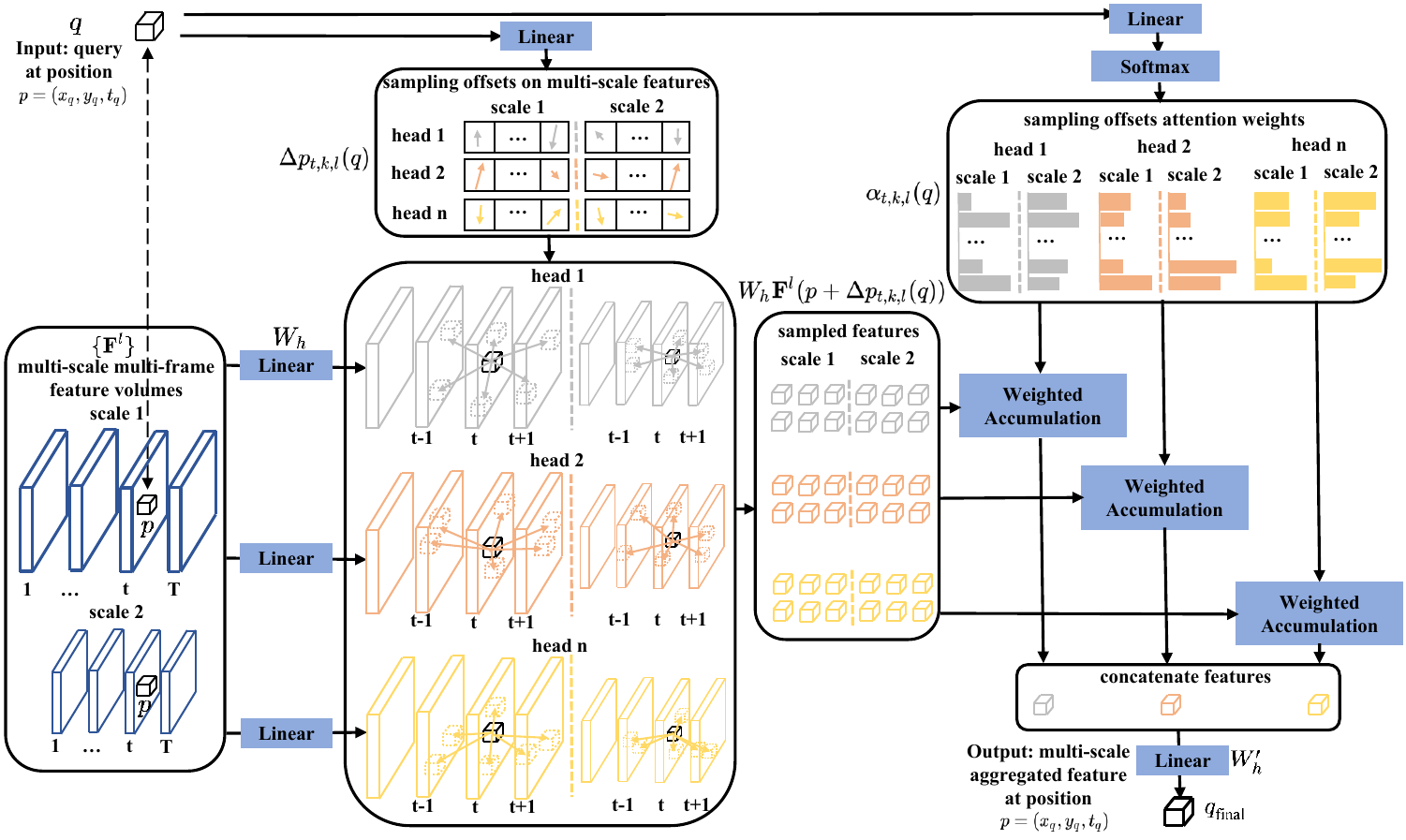}
    \caption{\textbf{Multi-scale Spatiotemporal Deformable Attention.} Given multi-scale (two scales as an example) multi-frame feature volumes, a query vector $q$, specified at position $(x_q, y_q, t_q)$, regresses multiple 2D offsets $\Delta p_{t,k,l}$ to sample features on a set of neighboring frames $\mathbf{S}^l(t_q)$ in \emph{all scales} of feature volume, as well as corresponding attention weights $\alpha_{t,k,l}$.
    The sampled features are aggregated and concatenated together to obtain the multi-scale aggregated feature as the output.
    }
    \label{fig:multiscale-attention}
\end{figure}

\boldstart{Extension to multi-scale features.} Our proposed spatiotemporal deformable attention mechanism can be naturally applied to multi-scale multi-frame features extracted by CNN backbone. The process is summarized in Fig.~\ref{fig:multiscale-attention}. For the query vector $q$ at the position $p=(x_q,y_q,t_q)$, we pass it through two linear layers to regress the sampling offsets $\Delta p_{t,k,l}$ and the attention weights $\alpha_{t,k,l}$ for all scales in parallel, where $l$ indexes the feature volume scale. 
We use the offsets $\Delta p_{t,k,l}$ to sample the image features at multiple scales and linearly combine these sampled features using the weights $\alpha_{t,k,l}$. 
This process is mathematically expressed as
\begin{equation}\small\begin{aligned}
    \label{eq:3dattention_ms}
    q_{\text{final}} &= \sum_{h} W'_{h} \Big[ \sum\limits_{t,k,l} \alpha_{t,k,l}(q)\cdot W_{h} \mathbf{F}^l\big(p+\Delta p_{t,k,l}(q)\big) \Big], 
\end{aligned}\end{equation}
where $W_h$ and $W'_h$ are the parameters of linear layers. 

The sampling at the higher scale focuses more on local features with relatively shorter field of perception, while at the lower scale the sampling obtains more global features with relatively broader field of perception. Unlike the self-attention where it is costly to attend to the feature volume globally and unable to extend to multi-scale features, our spatiotemporal deformable attention is efficient as it sparsely sample the image feature and approximates global attention by repeating the sparse sampling for several stages.

\begin{figure*}[ptb]
    \centering
    \includegraphics[width=\textwidth]{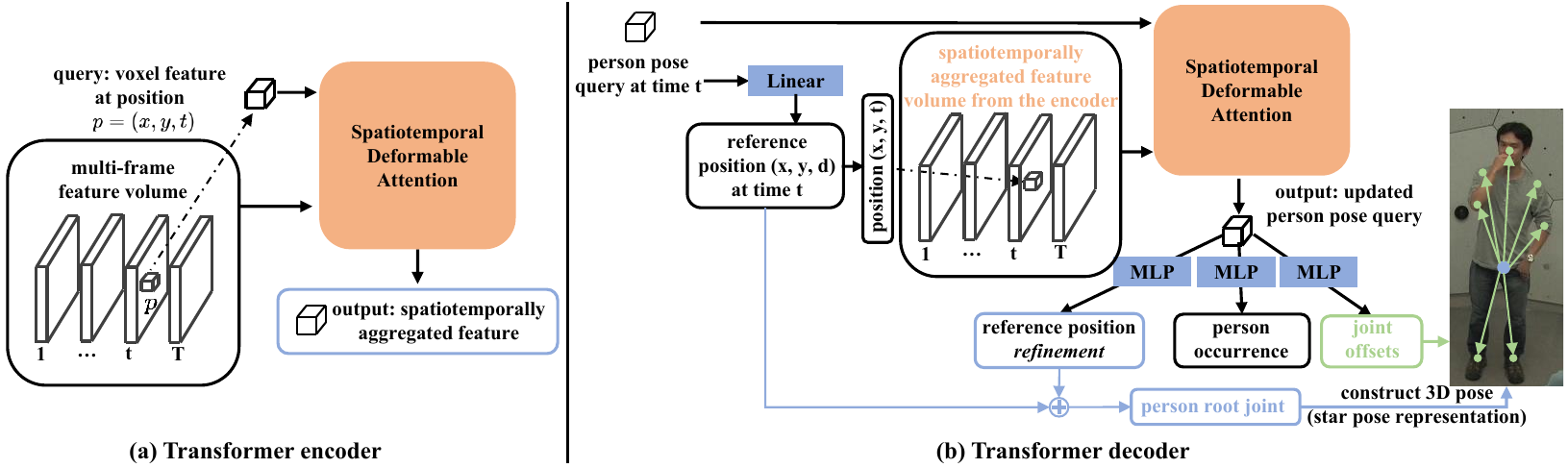}
    \caption{\textbf{Architecture of transformer encoder and decoder with spatiotemporal deformable attention}. (a) In the encoder, for the voxel at position $(x, y, t)$ of multi-frame feature volume $\mathbf{F}$, its voxel feature acts as the query in the attention module to aggregate spatiotemporal features from the feature volume. This process covers every voxel in the volume to create a spatiotemporally aggregated feature volume.
    (b) In the decoder, a learnable person pose query $q_t\in \mathbb{R}^C$ at time $t$ first regresses a reference position $(x, y, d)$, and then conducts spatiotemporal deformable attention at the position $(x, y, t)$ of the spatiotemporally aggregated feature volume to aggregate useful pose features. The updated person pose query passes through three MLPs to predict the refinement over reference position $(\Delta x, \Delta y, \Delta d)$, joint offsets $\mathbf{J}^{\text{offsets}}$ with joint visibility $\mathbf{V}$, and the person occurrence probability $o$ at time $t$. The refined reference position is regarded as person root joint and together with joint offsets and occurrence probability, the person's 3D pose $\mathbf{P}_t$ at time $t$ can be constructed.
    }
    \label{fig:encoder-decoder}
\end{figure*}

\subsection{Spatiotemporal Transformer Encoder}
\label{sec:encoder}
The goal of the transformer encoder is to generate spatiotemporally aggregated feature volume from the CNN-extracted multi-frame features. Fig.~\ref{fig:encoder-decoder}~(a) describes our transformer encoder with single layer of attention. The encoder consists of multiple layers of attention module where the refined feature volume is used as the input to the next layer and iteratively aggregate spatiotemporal features. Details on multi-layer attention design are given in the supplementary.

\boldstart{Spatiotemporal positional encoding.} The position encoding of the pixel location is essential to the transformer attention mechanism. Our encoding scheme follows Wang et al.~\cite{wang2021end} for a video snippet. Each location $(x, y, t)$ is independently applied to $C/3$ sine and cosine functions with different frequencies as in Vaswani et al.~\cite{vaswani2017attention} to generate encodings. These encodings are concatenated to form the final $C$ channel positional encoding, which are added to the feature volume $\mathbf{F}$ and fed to the transformer encoder.

\boldstart{Joint heatmap supervision.} It is commonly observed that pose estimation is better aligned with the input image if it is regressed from the joint heatmap or body part segmentation~\cite{varol2018bodynet}. Inspired by Habibie et al.~\cite{habibie2019wild}, we enforce the first $N_J$ channels of each temporal slice in the volume to be the multi-person joints heatmap, denoted by $\mathbf{H}_t$. We find empirically that this intermediate supervision improves the 3D pose accuracy (2.9\% of 3D-PCK).

\subsection{Spatiotemporal Transformer Decoder}
\label{sec:decoder} 
A person pose query $q\in\mathbb{R}^C$ in the decoder is a feature vector or embedding corresponding to a single person's pose at a specific time. Given a fixed number of $N(T + T_{f})$ learnable person pose queries, the decoder updates these queries by accumulating pose features from the spatiotemporally aggregated feature volume $\mathbf{F}$. These pose queries are used to regresses $N$ people's 3D pose trajectories in a single shot. Each person's trajectory consists of $T$ observed poses and $T_f$ future poses.

\boldstart{Temporal positional encoding.} Since the pose queries of each person are agnostic about the chronological order and need to predict a pose trajectory, we add ${T + T_f}$ learnable temporal positional encoding to each query to make it aware of its order before feeding to the transformer decoder. We empirically observe that this helps estimate more accurate 3D pose trajectory in supplementary. 

\boldstart{Pose querying.} The process is illustrated in Fig.~\ref{fig:encoder-decoder}~(b). Given a person pose query $q_t$ at time $t$, we first regress a reference position $(x, y, d)$ and then conduct spatiotemporal deformable attention at the position $(x, y, t)$ of $\mathbf{F}$. After updating person pose query, it is passed through 3 parallel MLPs to estimate the refinement over the reference position, person occurrence probability at time $t$, and joint offsets with joint visibility. The refined reference position is regarded as the root joint of the person, and we can construct the 3D pose $\mathbf{P}_t$ at time $t$ together with the predicted joint offsets and occurrence probability. Similarly, the decoder stacks multiple layers of attention module to iteratively update pose query with detailed design described in the supplementary.

\subsection{Trajectory Matching Loss}
\label{sec:losses}
Snipper predicts a fixed number of $N$ people trajectories within the snippet in a single shot, where each trajectory can be represented as $\mathbf{\Gamma}_i = \{\mathbf{P}_t^{(i)}\}_{t=1}^{T + T_f}$. To supervise Snipper, we use Hungarian algorithm~\cite{kuhn1955hungarian} to find the optimal matches between the predicted and target pose trajectories, and form the pose trajectory loss for back propagation.

\boldstart{Hungarian matching cost}. Let $\mathbf{\Gamma}= \{ \mathbf{\Gamma}_i \}_{i=1}^N$ and $\hat{\mathbf{\Gamma}}= \{ 
\hat{\mathbf{\Gamma}}_{i} \}_{i=1}^M$ be the predicted and target pose trajectory sets, respectively. We use Hungarian algorithm to find an optimal permutation $\hat \sigma$ of $\mathbf{\Gamma}$ with the lowest bipartite matching cost,
\begin{equation} \small
    \label{eq:matching-cost} \begin{aligned}
   \hat\sigma = \mathop{\arg\min}_{\sigma} \sum_{i=1}^M &\mathcal{L}_{\text{occ}}(\mathbf{\Gamma}_{\sigma_i}, \hat{\mathbf{\Gamma}}_i) + \mathcal{L}_{\text{traj}}(\mathbf{\Gamma}_{\sigma_i}, \hat{\mathbf{\Gamma}}_i) + \mathcal{L}_{\text{vis}}(\mathbf{\Gamma}_{\sigma_i}, \hat{\mathbf{\Gamma}}_i),
\end{aligned}\end{equation}
where $\mathcal{L}_{\text{occ}}(\mathbf{\Gamma}_{\sigma_i}, \hat{\mathbf{\Gamma}}_i)$ is the negative average probability of occurrence,
\begin{equation} \small
    \mathcal{L}_{\text{occ}}(\mathbf{\Gamma}_{\sigma_i}, \hat{\mathbf{\Gamma}}_i)=-\frac{\sum_{t} \mathds{1}(\hat o^{(i)}_{t} \neq \emptyset) \cdot o^{(\sigma_i)}_{t}}{\sum_{t} \mathds{1}(\hat o^{(i)}_{t} \neq \emptyset)},
\end{equation}
where $\hat o^{(i)}_{t} \neq \emptyset$ means the $i$-th target person occurs at time $t$, $\mathcal{L}_{\text{traj}}(\mathbf{\Gamma}_{\sigma_i}, \hat{\mathbf{\Gamma}}_i)$ measures the average $L1$ distance between the predicted and visible target pose trajectory,
\begin{equation} \small
    \label{eq:traj_cost}
    \mathcal{L}_{\text{traj}}(\mathbf{\Gamma}_{\sigma_i}, \hat{\mathbf{\Gamma}}_i) = \frac{\sum_k\sum_t  \| \hat{V}^{(i)}_{k,t} \cdot (J^{(\sigma_i)}_{k,t} - \hat{J}^{(i)}_{k,t})\|_1}{\sum_k\sum_t \hat{V}^{(i)}_{k,t} },
\end{equation}
and $\mathcal{L}_{\text{vis}}(\mathbf{\Gamma}_{\sigma_i}, \hat{\mathbf{\Gamma}}_i)$ is the average $L2$ distance between the predicted and target joint visibility,
\begin{equation} \small
    \label{eq:vis_cost}
    \mathcal{L}_{\text{vis}}(\mathbf{\Gamma}_{\sigma_i}, \hat{\mathbf{\Gamma}}_i) = \frac{\sum_{k}\sum_{t} \| V^{(\sigma_i)}_{k,t} - \hat{V}^{(i)}_{k,t}\|_2^2}{N_j(T+T_f)} .
\end{equation}
In the above cost definition, we simplify the notations with 
$\sum_t$ as the iteration over all the time steps $\{1,\dots,T+T_f\}$, and $\sum_k$ as the iteration over all the joints $\{1,\dots,N_J\}$ respectively. We follow Carion et al.~\cite{carion2020end} and adopt the detection probability instead of the log-probabilities in $\mathcal{L}_{\text{occ}}(\mathbf{\Gamma}_{\sigma_i}, \hat{\mathbf{\Gamma}}_i)$. We observe better matching behavior between the predicted and target pose trajectories especially for earlier epochs with this strategy.

\boldstart{Training loss.} Given the optimal permutation $\hat{\sigma}$, the matched predictions are used to compute both person occurrence and 3D pose losses, and the remaining predictions are only used to compute person occurrence loss. We define the total training loss as 
\begin{equation} \small \begin{aligned}
    \label{eq:hungarian-loss}
    \mathcal{L}_{\text{train}} = &\sum_{i=1}^{M} \Big( \mathcal{L}_{\text{occ}}'(\mathbf{\Gamma}_{\hat{\sigma}_i}, \hat{\mathbf{\Gamma}}_i) +\mathcal{L}_{\text{traj}}(\mathbf{\Gamma}_{\hat{\sigma}_i}, \hat{\mathbf{\Gamma}}_i) + \mathcal{L}_{\text{vis}}(\mathbf{\Gamma}_{\hat{\sigma}_i}, \hat{\mathbf{\Gamma}}_i) \\ &+ \mathcal{L}_{\text{offset}}(\mathbf{\Gamma}_{\hat{\sigma}_i}, \hat{\mathbf{\Gamma}}_i) + \mathcal{L}_{\text{smooth}}(\mathbf{\Gamma}_{\hat{\sigma}_i}, \hat{\mathbf{\Gamma}}_i) \Big) + 
    \mathcal{L}_{\text{heatmap}}(\mathbf{H}, \mathbf{\hat H}),
\end{aligned} \end{equation}
where $\mathcal{L}'_{\text{occ}}$ is the negative log-likelihood for person occurrence prediction, 
\begin{align}\small
    \label{eq:trainOcc}
    \mathcal{L}_{\text{occ}}'(\mathbf{\Gamma}_{\hat{\sigma}_i}, \hat{\mathbf{\Gamma}}_i) = - \sum_t \mathds{1}(\hat{o}^{(i)}_{t} \neq \emptyset) \cdot \log o^{(\hat{\sigma}_i)}_{t}.
\end{align}
$\mathcal{L}_{\text{traj}}(\mathbf{\Gamma}_{\hat \sigma_i}, \hat{\mathbf{\Gamma}}_i)$ and $\mathcal{L}_{\text{vis}}(\mathbf{\Gamma}_{\hat \sigma_i}, \hat{\mathbf{\Gamma}}_i)$ are defined in Eq.~\ref{eq:traj_cost} and~\ref{eq:vis_cost} with the permutation replaced with the optimal one $\hat \sigma_i$ for the computation of training loss.

For the following losses, we drop the superscript $\hat{\sigma}_i$ and $i$ for $J^{\text{offset}}$ and $\hat{J}^{\text{offset}}$ for simplicity. $\mathcal{L}_{\text{offset}}(\mathbf{\Gamma}_{\hat{\sigma}_i}, \hat{\mathbf{\Gamma}}_i)$ measures the average $L1$ distance between the predicted and target visible joint offsets for the supervision of a single person's pose,
\begin{equation} \small
    \label{eq:offset_cost}
    \mathcal{L}_{\text{offset}}(\mathbf{\Gamma}_{\hat{\sigma}_i}, \hat{\mathbf{\Gamma}}_i) = \frac{\sum_k\sum_t  \| \hat{V}_{k,t} \cdot (J_{k,t}^{\text{offset}} - \hat{J}_{k,t}^{\text{offset}}) \|_1}{\sum_k\sum_t \hat{V}_{k,t} },
\end{equation}
$\mathcal{L}_{\text{smooth}}(\mathbf{\Gamma}_{\hat{\sigma}_i}, \hat{\mathbf{\Gamma}}_i)$ is the average $L2$ smoothness of joint offsets between frames within a video snippet,
\begin{equation} \small
    \label{eq:smooth_cost}
    \mathcal{L}_{\text{smooth}}(\mathbf{\Gamma}_{\hat{\sigma}_i}, \hat{\mathbf{\Gamma}}_i) = \frac{\sum_k\sum_t \| J_{k,t}^{\text{offset}} - J_{k,t-1}^{\text{offset}} \|_2^2}{N_j(T+T_f-1)},
\end{equation}
and $\mathcal{L}_{\text{heatmap}}$ is the average $L2$ distance of joints heatmaps,
\begin{equation} \small
    \label{eq:heatmap_cost}
    \mathcal{L}_{\text{heatmap}} = \frac{1}{T}\sum_t \| \mathbf{H}_{t} - \mathbf{\hat H}_{t} \|_2^2.
\end{equation}

Note that while Eq.~\ref{eq:traj_cost} already captures Eq.~\ref{eq:offset_cost} to some extents, Eq.~\ref{eq:traj_cost} couples the root and the offsets, and thus hurts learning. We empirically observe that adding Eq.~\ref{eq:offset_cost} leads to faster convergence. As the camera motion at each frame is unknown and our predicted root joint is relative to the camera coordinate, $\mathcal{L}_{\text{smooth}}$ factors out the root motion and ensures smooth joints motion.

We apply intermediate pose supervision by computing the losses in Eq.~\ref{eq:hungarian-loss}, except for the heatmap loss, for each layer of the decoder to guide the learning. Besides, we normalize these losses by the number of target trajectories within a batch such that its magnitude is approximately the same across batches.

\section{Experiments}
We evaluate on three datasets: JTA~\cite{fabbri2018learning}, CMU-Panoptic~\cite{joo2017panoptic} and Posetrack2018~\cite{PoseTrack}.

\boldstart{Evaluation Metrics.} Our method involves three tasks, multi-person pose estimation, tracking and motion forecasting. For \textit{3D pose estimation}, MPJPE is used to evaluate 3D pose accuracy in mm, and MPJPE$^\text{rel}$ means MPJPE after root joint alignment between predicted and target pose. To compare with~\cite{benzine2020pandanet}, we report 3D-PCK, where a joint is considered as correct if the distance from the corresponding ground truth joint is less than $150$mm. To compare with~\cite{cheng2021monocular,fabbri2020compressed} on JTA dataset, we also report F1 $@thr\in\{0.4, 0.8, 1.2\}$ meters, where a joint is considered as correct if the joint position error is less than $thr$. To compare with~\cite{openpose,fabbri2018learning,wang2020combining,yu2018multi,hwang2019pose,raaj2019efficient}, we report the AP defined in~\cite{PoseTrack} on JTA and Posetrack datasets. 
For \textit{tracking}, we follow~\cite{PoseTrack,reddy2021tessetrack} to report MOTA metrics defined either on 3D or 2D pose. 
For \textit{motion forecasting}, we report MPJPE$^\text{rel}$ for 3D pose estimation and 3D path error of root joint, following~\cite{cao2020long}, in Tab.~\ref{tab:JTA_future_motion}.

\boldstart{Implementation.} We use ResNet50 as the backbone to extract multi-scale features from each image and stack them into a feature volume $\mathbf{F}^l$ in chronological order, where $l=\{3, 4, 5\}$ indexes the convolution stage of feature map. The multi-scale feature volumes are transformed by a $1\times 1$ convolution to be $C=384$ channels.
In Snipper, 6 transformer encoder and decoder layers are used with 8 heads in each deformable attention module at the center frame and are halved every frame away from that frame. The head is initialized to have identical attention weights with initial offsets uniformly distributed in angle directions from starting 0 degree. 
We train Snipper on 8 $V100$ GPUs with batch size of 16 on JTA~\cite{fabbri2018learning} at 6FPS, CMU-Panoptic~\cite{joo2017panoptic} at 3FPS Posetrack2018~\cite{PoseTrack} at 7.5 FPS, and COCO, where we apply 2D transformation to make it become a video snippet, jointly.
We use 14 joints format in  MPII~\cite{andriluka20142d} common across all datasets.

\begin{table*}[ptb]
    \centering
    \caption{Quantitative evaluations of 3D pose tracking on JTA dataset. We compare with three bottom-up (BU) methods~\cite{openpose,fabbri2018learning,fabbri2020compressed}, a single-stage (SS) method~\cite{benzine2020pandanet}, and a hybrid of bottom-up and top-down (TD) method~\cite{cheng2021monocular}. Ours~(${\scriptstyle T=1}$) and Ours~(${\scriptstyle T=4}$) are the evaluations of model trained on a snippet of 1 and 4 frames. Ours~(${\scriptstyle t=4+1}$) and Ours~(${\scriptstyle t=4+2}$) are the evaluations of motion forecasting for the 1st and 2nd future frame based on the observed snippet of 4 frames. Snipper outperforms prior works on both tasks. Underline means the second best results.}
    \setlength{\tabcolsep}{1.5mm}
    \resizebox{0.8\textwidth}{!}{
    \begin{tabular}{@{}c|c|p{35pt}p{40pt}p{40pt}p{40pt}p{40pt}|p{40pt}@{}}
    \bottomrule \hline
        \multicolumn{2}{c|}{\multirow{2}{*}{Method}} & \multicolumn{5}{c|}{Pose Estimation} & \makecell[c]{Tracking} \\
    \cline {3-8}
         \multicolumn{2}{c|}{} & \makecell[c]{AP} & \makecell[c]{F1$@0.4$m} & \makecell[c]{F1$@0.8$m} & \makecell[c]{F1$@1.2$m} & \makecell[c]{3D-PCK} & \makecell[c]{MOTA} \\
    \hline
        \makecell[c]{OpenPose(2019)~\cite{openpose}} & BU & \makecell[c]{50.1} & \makecell[c]{-} & \makecell[c]{-} & \makecell[c]{-}  & \makecell[c]{-} & \makecell[c]{-}\\ 
        \makecell[c]{THOPA(2019)~\cite{fabbri2018learning}} & BU & \makecell[c]{59.3} & \makecell[c]{-} & \makecell[c]{-} & \makecell[c]{-}  & \makecell[c]{-} & \makecell[c]{59.3}\\
        \makecell[c]{LoCOn(2020)~\cite{fabbri2020compressed}} & BU & \makecell[c]{-} & \makecell[c]{50.8} & \makecell[c]{64.8}  & \makecell[c]{70.4} & \makecell[c]{-} & \makecell[c]{-}\\
        \makecell[c]{PandaNet(2020)~\cite{benzine2020pandanet}} & SS & \makecell[c]{-} & \makecell[c]{-} & \makecell[c]{-}  & \makecell[c]{-} & \makecell[c]{83.2} & \makecell[c]{-}\\
        \makecell[c]{Cheng et al.(2021)~\cite{cheng2021monocular}} & BU+TD & \makecell[c]{-} & \makecell[c]{57.2} & \makecell[c]{68.5} & \makecell[c]{72.9}  & \makecell[c]{-} & \makecell[c]{-}\\
    \hline
        \makecell[c]{Ours~(${\scriptstyle t=4+1}$)} & SS & \makecell[c]{66.5} & \makecell[c]{56.2} & \makecell[c]{67.9} & \makecell[c]{73.1}  & \makecell[c]{83.8} & \makecell[c]{-}\\
        
        \makecell[c]{Ours~(${\scriptstyle t=4+2}$)} & SS & \makecell[c]{64.5} & \makecell[c]{53.2} & \makecell[c]{65.9} & \makecell[c]{71.2}  & \makecell[c]{82.8} & \makecell[c]{-}\\
    \hline
        \makecell[c]{Ours~(${\scriptstyle T=1}$)} & SS & \makecell[c]{\underline{65.3}} & \makecell[c]{\underline{59.7}} & \makecell[c]{\underline{70.7}} & \makecell[c]{\underline{75.7}}  & \makecell[c]{\underline{83.4}} & \makecell[c]{\underline{61.4}}\\
        \makecell[c]{Ours~(${\scriptstyle T=4}$)} & SS & \makecell[c]{\textbf{70.5}} & \makecell[c]{\textbf{60.3}} & \makecell[c]{\textbf{71.5}} & \makecell[c]{\textbf{76.4}}  & \makecell[c]{\textbf{85.7}} & \makecell[c]{\textbf{63.2}}\\
    \hline \toprule  
    \end{tabular}
    }
    \label{tab:jta-evaluation}
\end{table*}

\begin{figure}[]
    \centering
    \includegraphics[width=0.8\columnwidth]{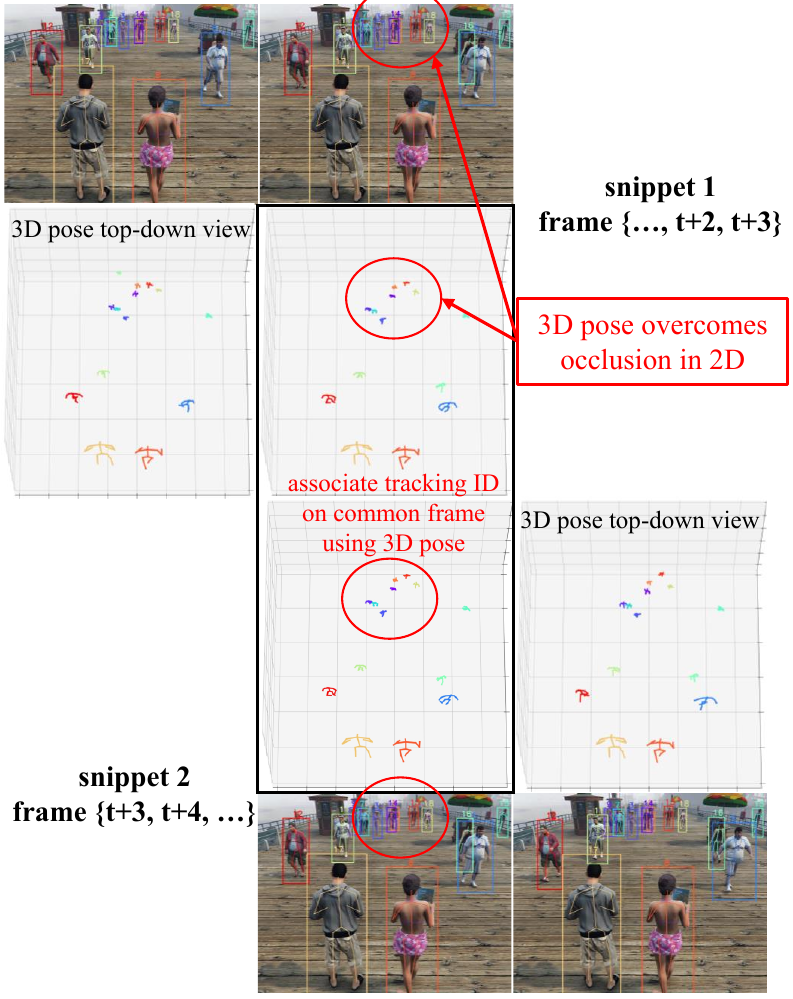}
    \caption{\textbf{Association between Two Consecutive Snippets.} To achieve multi-person tracking over the whole video, for two consecutive snippets (${\scriptstyle T=4}$) consisting of frames $\scriptstyle \{t,...,t+3\}$ and $\scriptstyle \{t+3,...,t+6\}$, the association of 3D pose tracking is based on the common frame $\scriptstyle t+3$. The 3D pose association helps overcome the occlusion on 2D images.}
    \label{fig:association}
\end{figure}

Ours~(${\scriptstyle T=1}$) and Ours~(${\scriptstyle T=4}$) denote the evaluation of model trained on a snippet of 1 and 4 frames.
For fair comparison in each dataset, we train and test our model on the corresponding dataset only following prior work.
To achieve \emph{multi-person tracking over the whole video}, for two consecutive snippets (${\scriptstyle T=4}$) consisting of frames $\{t,...,t+3\}$ and $\{t+3,...,t+6\}$, the association of 3D pose tracking is based on the common frame $t+3$ with the nearest 3D pose matching measured in Euclidean distance. The process is shown in Fig.~\ref{fig:association}.
For snippet (${\scriptstyle T=1}$), whole-video tracking is achieved by Hungarian matching on poses of two consecutive frames. For motion forecasting, ${\scriptstyle T_f=2}$ is used based on ${\scriptstyle T=4}$ observed frames. Ours~(${\scriptstyle t=4+1}$) and Ours~(${\scriptstyle t=4+2}$) denote the evaluation on predicted pose of the 1st and 2nd future frame. The inference time is 76ms and 266ms in average for a single snippet of 1 and 4 frames on JTA dataset on a single $V100$ GPU, and the number of parameters is 40M and 43M respectively.

\subsection{JTA Evaluation}
For JTA dataset, we resize the input image to the resolution $540\times 960$ and downsample the video to 6 FPS. Since there is no prior work evaluating all the three tasks on JTA dataset, we present the results from our method in Tab.~\ref{tab:jta-evaluation} and compare with the state-of-art methods on multi-person 3D pose estimation, tracking and motion forecasting. For pose estimation, comparing with the single-stage PanadaNet~\cite{benzine2020pandanet}, our method estimates 0.2\% and 2.5\% better 3D-PCK for Ours~(${\scriptstyle T=1}$) and Ours~(${\scriptstyle T=4}$), respectively. Compared with~\cite{fabbri2020compressed,cheng2021monocular}, Ours~(${\scriptstyle T=4}$) shows around 10\% and 3\% of F1$@0.4$m increase. For tracking, Ours~(${\scriptstyle T=4}$) is around 4\% more accurate in MOTA than THOPA~\cite{fabbri2018learning}. For motion forecasting, Ours~(${\scriptstyle T=4}$) can predict competitive future motion that outperforms methods that do prediction on the hidden future images ~\cite{openpose,fabbri2018learning,chen2018cascaded}, even with a single-stage architecture (\eg, 66.5, 64.5 of F1@0.4m and 83.8, 82.8 of 3D-PCK for the 2 future frames).
In Tab.~\ref{tab:JTA_future_motion}, we compare our method with HMP~\cite{cao2020long}, the only work to forecast 3D pose from RGB images, on motion forecasting of next 2 frames. Our method estimates comparable pose forecasting as HMP but outperforms it when noise is added to the history pose in HMP, highlighting the benefit of solving the pose estimation, tracking and forecasting in a joint framework. These accuracy improvement is rooted in the effectiveness of the spatiotemporal deformable attention.

The evaluation of motion forecasting on JTA dataset is presented in Tab.~\ref{tab:JTA_future_motion}. No forecasting means to keep the last observed pose for evaluation without motion prediction. For fair comparison, we retrain HMP~\cite{cao2020long} to take only 4 frames as input and forecast the next 2 frames. The evaluation is done for the deterministic mode of HMP. We use the ground truth history 2D pose in $\text{HMP}^1$ but added Gaussian noise $\mathcal{N}(0,3)$ pixels to the ground truth history 2D pose for $\text{HMP}^2$. HMP (Hourglass), means HMP is trained with 2D poses estimated by Hourglass~\cite{newell2016stacked}. Our method jointly estimates the poses in the observed frames and forecasts its motion, which shows comparable accuracy with noise-free HMP but noticeably outperforms it when adding noise to the history pose or using the estimated poses.

\begin{table}[ptb]
    \centering
    \caption{Quantitative evaluations of motion forecasting on JTA dataset. Our method shows comparable accuracy with noise-free HMP but noticeably outperforms it when adding noise to the history pose or using the estimated poses}
    \setlength{\tabcolsep}{3.5mm}
    \resizebox{\columnwidth}{!}{
    \begin{tabular}{@{}c|cc|cc@{}}
    \bottomrule \hline
        \makecell[c]{Method} & \multicolumn{2}{c|}{3D Path Error (mm)} & \multicolumn{2}{c}{MPJPE$^\text{rel}$ (mm)} \\
    \hline
        \makecell[c]{Forecasting Time} & \makecell[c]{166ms} & \makecell[c]{333ms} & \makecell[c]{166ms} & \makecell[c]{333ms} \\ 
    \hline
        \makecell[c]{No forecasting} & \makecell[c]{353.5} & \makecell[c]{409.1} & \makecell[c]{123.5} & \makecell[c]{139.1} \\ 
        \makecell[c]{$\text{HMP}^1$~\cite{cao2020long}} & \makecell[c]{90.3} & \makecell[c]{112.6} & \makecell[c]{35.4} & \makecell[c]{39.5} \\ 
        \makecell[c]{$\text{HMP}^2$~\cite{cao2020long}} & \makecell[c]{94.5} & \makecell[c]{121.8} & \makecell[c]{48.5} & \makecell[c]{61.4} \\ 
        \makecell[c]{$\text{HMP}$~(Hourglass~\cite{newell2016stacked})} & \makecell[c]{95.2} & \makecell[c]{123.3} & \makecell[c]{46.8} & \makecell[c]{60.6} \\ 
    \hline
        \makecell[c]{Ours} & \makecell[c]{92.3} & \makecell[c]{117.7} & \makecell[c]{37.9} & \makecell[c]{43.0}\\
    \hline \toprule  
    \end{tabular}
    }
    \label{tab:JTA_future_motion}
\end{table}

\subsection{CMU-Panoptic Evaluation}
For multi-person pose estimation or tracking, there are mainly 2 protocols of data split used in prior works~\cite{benzine2020pandanet,reddy2021tessetrack}. Protocol 1 follows~\cite{reddy2021tessetrack} where 3 views HD cameras ($3, 13, 23$) and all the haggling videos of version 1.2 are used. The training and testing video split follows~\cite{reddy2021tessetrack,joo2019towards}, and the evaluation metrics follow~\cite{reddy2021tessetrack,tu2020voxelpose}. Protocol 2 follows~\cite{benzine2020pandanet,fabbri2020compressed} where 4 scenarios (Haggling, Mafia, Ultimatum, Pizza) are selected. The testing set is composed of HD videos of camera 16 and 30, and the training set includes videos of other 28 cameras. The evaluation metrics follow~\cite{benzine2020pandanet,fabbri2020compressed}. Since the speed motion of this dataset is slower than JTA dataset, we downsample to 3 FPS to avoid too much redundancy during training and use the image resolution $540\times960$ as the input. We also provide test results of 6 FPS with the model trained on 3 FPS, denoted by Ours~(${\scriptstyle T=4}$, 6 FPS).

\begin{table}
    \centering
    \caption{Quantitative evaluations on CMU-Panoptic of \textbf{protocol 1}. Our method shows competitive or higher accuracy than three top-down approaches~\cite{tu2020voxelpose,reddy2021tessetrack,zhang2022voxeltrack}, and gives competitive accuracy of motion forecasting. Underline means the second best results.}
    \setlength{\tabcolsep}{2.5mm}
    \resizebox{\columnwidth}{!}{
    \begin{tabular}{@{}c|c|c|ccc@{}}
    \bottomrule \hline
        \multicolumn{2}{c|}{Method} & \makecell[c]{Backbone} & \makecell[c]{MPJPE} & \makecell[c]{MPJPE$^\text{rel}$} & \makecell[c]{MOTA} \\
    \hline
        \makecell[c]{VoxelPose(2020)~\cite{tu2020voxelpose}} & TD & \makecell[c]{\footnotesize ResNet50} & \makecell[c]{66.9} & \makecell[c]{51.1} & \makecell[c]{-} \\ 
        \makecell[c]{TesseTrack(2021)~\cite{reddy2021tessetrack}} & TD & \makecell[c]{\footnotesize HRNet} & \makecell[c]{\textbf{18.9}} & \makecell[c]{-} & \makecell[c]{76.0} \\
        \makecell[c]{VoxelTrack(2022)~\cite{zhang2022voxeltrack}} & TD & \makecell[c]{\footnotesize DLA-34} & \makecell[c]{66.4} & \makecell[c]{-} & \makecell[c]{-} \\
    \hline
        \makecell[c]{Ours~(${\scriptstyle t=4+1}$)} & SS & \makecell[c]{\footnotesize ResNet50} & \makecell[c]{49.0} & \makecell[c]{40.8} & \makecell[c]{-} \\
        \makecell[c]{Ours~(${\scriptstyle t=4+2}$)} & SS & \makecell[c]{\footnotesize ResNet50} & \makecell[c]{50.7} & \makecell[c]{41.3} & \makecell[c]{-} \\
        \makecell[c]{Ours~(${\scriptstyle T=4}$, 6 FPS)} & SS & \makecell[c]{\footnotesize ResNet50} & \makecell[c]{45.1} & \makecell[c]{37.3} & \makecell[c]{80.9} \\
    \hline
        \makecell[c]{Ours~(${\scriptstyle T=1}$)} & SS & \makecell[c]{\footnotesize ResNet50} & \makecell[c]{48.4} & \makecell[c]{\underline{37.5}} & \makecell[c]{\underline{78.1}} \\
        \makecell[c]{Ours~(${\scriptstyle T=4}$)} & SS & \makecell[c]{\footnotesize ResNet50} & \makecell[c]{\underline{44.3}} & \makecell[c]{\textbf{37.1}} & \makecell[c]{\textbf{81.7}} \\
    \hline \toprule  
    \end{tabular}
    }
    \label{tab:panoptic-evaluation1}
\end{table}

\begin{table}
    \centering
    \caption{Quantitative evaluations on CMU-Panoptic of \textbf{protocol 2}. Our single-stage (SS) method outperforms two bottom-up~\cite{zanfir2018deep,fabbri2020compressed} and three single-stage methods~\cite{benzine2020pandanet,wang2022distribution,jin2022single,benzine2021single} and gives competitive accuracy of motion forecasting.}
    \label{tab:panoptic-evaluation2}
    \setlength{\tabcolsep}{1.5mm}
    \resizebox{\columnwidth}{!}{
    \begin{tabular}{@{}c|c|ccccc|c|c@{}}
    \bottomrule \hline
        \multicolumn{2}{c|}{\multirow{2}{*}{Method}} & \multicolumn{5}{c|}{MPJPE} &  \multirow{2}{*}{F1} &  \multirow{2}{*}{MOTA}\\
    \cline{3-7}
         \multicolumn{2}{c|}{} & \makecell[c]{Hag.} & \makecell[c]{Maf.} & \makecell[c]{Ult.} & \makecell[c]{Piz.} & \makecell[c]{Avg.} & \makecell[c]{} & \makecell[c]{} \\
    \hline
        \makecell[c]{MubyNet(2018)~\cite{zanfir2018deep}} & BU & \makecell[c]{72.4} & \makecell[c]{78.8} & \makecell[c]{66.8} & \makecell[c]{94.3} & \makecell[c]{72.1} & \makecell[c]{-} & \makecell[c]{-}\\ 
        \makecell[c]{LoCO(2020)~\cite{fabbri2020compressed}} & BU & \makecell[c]{45} & \makecell[c]{95} & \makecell[c]{58} & \makecell[c]{79} & \makecell[c]{69} & \makecell[c]{89.2} & \makecell[c]{-}\\ 
        \makecell[c]{PandaNet(2020)~\cite{benzine2020pandanet}} & SS & \makecell[c]{40.6} & \makecell[c]{37.6} & \makecell[c]{\textbf{31.3}} & \makecell[c]{55.8} & \makecell[c]{42.7} & \makecell[c]{-} & \makecell[c]{-}\\
        \makecell[c]{Benzine et al.(2021)~\cite{benzine2021single}} & SS & \makecell[c]{70.1} & \makecell[c]{66.6} & \makecell[c]{55.6} & \makecell[c]{78.4} & \makecell[c]{68.5} & \makecell[c]{-} & \makecell[c]{-}\\
        \makecell[c]{Jin et al.(2022)~\cite{jin2022single}} & SS & \makecell[c]{63.7} & \makecell[c]{58.5} & \makecell[c]{52.3} & \makecell[c]{69.1} & \makecell[c]{60.9} & \makecell[c]{-} & \makecell[c]{-}\\
        \makecell[c]{Wang et al.(2022)~\cite{wang2022distribution}} & SS & \makecell[c]{53.3} & \makecell[c]{51.2} & \makecell[c]{49.1} & \makecell[c]{61.5} & \makecell[c]{53.8} & \makecell[c]{-} & \makecell[c]{-}\\
    \hline
        \makecell[c]{Ours~(${\scriptstyle t=4+1}$)} & SS & \makecell[c]{41.4} & \makecell[c]{38.8} & \makecell[c]{41.6} & \makecell[c]{44.9} & \makecell[c]{40.3} & \makecell[c]{88.7} & \makecell[c]{-}\\
        \makecell[c]{Ours~(${\scriptstyle t=4+2}$)} & SS & \makecell[c]{43.0} & \makecell[c]{40.9} & \makecell[c]{42.9} & \makecell[c]{47.4} & \makecell[c]{42.4} & \makecell[c]{85.5} & \makecell[c]{-}\\
        \makecell[c]{Ours~(${\scriptstyle T=4}$, 6 FPS)} & SS & \makecell[c]{37.3} & \makecell[c]{37.1} & \makecell[c]{39.0} & \makecell[c]{42.6} & \makecell[c]{38.2} & \makecell[c]{90.0} & \makecell[c]{93.0} \\
    \hline
        \makecell[c]{Ours~(${\scriptstyle T=1}$)} & SS & \makecell[c]{\underline{37.6}} & \makecell[c]{\underline{38.5}} & \makecell[c]{39.7} & \makecell[c]{\underline{45.0}} & \makecell[c]{\underline{39.4}} & \makecell[c]{\underline{89.4}} & \makecell[c]{\underline{92.9}}\\
        \makecell[c]{Ours~(${\scriptstyle T=4}$)} & SS & \makecell[c]{\textbf{36.8}} & \makecell[c]{\textbf{36.9}} & \makecell[c]{\underline{38.6}} & \makecell[c]{\textbf{42.5}} & \makecell[c]{\textbf{37.9}} & \makecell[c]{\textbf{90.1}} & \makecell[c]{\textbf{93.4}}\\
    \hline \toprule  
    \end{tabular}
    }
\end{table}

The results of protocol 1 are shown in Tab.~\ref{tab:panoptic-evaluation1}. Our method outperforms VoxelPose~\cite{tu2020voxelpose}: 22.6mm (+33\%) on MPJPE and 14mm (+27\%) on relative MPJPE. As for its following work VoxelTrack~\cite{zhang2022voxeltrack}, we also exceeds 22.1mm on MPJPE. Compare with TesseTrack~\cite{reddy2021tessetrack}, Ours~(${\scriptstyle T=4}$) shows higher pose tracking accuracy (81.7 vs 76.0 of MOTA), but lower accuracy on MPJPE, which might be because TesseTrack uses HRNet as the backbone (around 100M parameters) while ours only uses ResNet50 (43M parameters).  For protocol 2, we show comparison with six recent works~\cite{zanfir2018deep,fabbri2020compressed,benzine2020pandanet,benzine2021single,jin2022single,wang2022distribution} in Tab.~\ref{tab:panoptic-evaluation2}. Snipper performs better on F1 scores and MPJPE in all six sequences except for the Ultimatum sequence, where PandaNet~\cite{benzine2020pandanet} achieves 7.3mm lower than ours. For tracking, Snipper achieves over 90\% MOTA. For motion prediction, ours~(${\scriptstyle t=T+1}$) and ours~(${\scriptstyle t=T+2}$) in both protocol 1 and 2 have competitive results on MPJPE, only about 3 and 5mm of worse than ours~(${\scriptstyle T=4}$) with the observed motion (see Tab.~\ref{tab:panoptic-evaluation2}).

\subsection{Posetrack2018 Evaluation}
\begin{table}[ptb]
    \centering
    \caption{Quantitative evaluations (AP) of pose estimation on Posetrack2018 val set.}
    \setlength{\tabcolsep}{1.1mm}
    \resizebox{\columnwidth}{!}{
    \begin{tabular}{@{}c|c|cccccccc@{}}
    \bottomrule \hline
        \multicolumn{2}{c|}{Method} & \makecell[c]{Head} & \makecell[c]{Sho} & \makecell[c]{Elb} & \makecell[c]{Wri} & \makecell[c]{Hip} & \makecell[c]{Kne} & \makecell[c]{Ank} & \makecell[c]{Avg}\\
    \hline
        \makecell[c]{DetTrack(2020)~\cite{wang2020combining}} & \makecell[c]{TD} & \makecell[c]{84.9} & \makecell[c]{87.4} & \makecell[c]{84.8} & \makecell[c]{79.2} & \makecell[c]{77.6} & \makecell[c]{79.7} & \makecell[c]{75.3} & \makecell[c]{\textbf{81.5}} \\
        \makecell[c]{PT\_CPN++(2018)~\cite{yu2018multi}} & \makecell[c]{TD} & \makecell[c]{82.4} & \makecell[c]{\textbf{88.8}} & \makecell[c]{\textbf{86.2}} & \makecell[c]{\textbf{79.4}} & \makecell[c]{72.0} & \makecell[c]{\textbf{80.6}} & \makecell[c]{\textbf{76.2}} & \makecell[c]{80.9} \\
    \hline
        \makecell[c]{TML++(2019)~\cite{hwang2019pose}} & \makecell[c]{BU} & \makecell[c]{-} & \makecell[c]{-} & \makecell[c]{-} & \makecell[c]{-} & \makecell[c]{-} & \makecell[c]{-} & \makecell[c]{-} & \makecell[c]{74.6} \\
        \makecell[c]{STAF(2019)~\cite{raaj2019efficient}} & \makecell[c]{BU} & \makecell[c]{-} & \makecell[c]{-} & \makecell[c]{-} & \makecell[c]{64.7} & \makecell[c]{-} & \makecell[c]{-} & \makecell[c]{62.0} & \makecell[c]{70.4} \\
    \hline
        \makecell[c]{Ours~(${\scriptstyle T=1}$)} & \makecell[c]{SS} & \makecell[c]{86.5} & \makecell[c]{85.6} & \makecell[c]{71.5} & \makecell[c]{67.9} & \makecell[c]{78.1} & \makecell[c]{72.0} & \makecell[c]{62.6} & \makecell[c]{74.9} \\
        \makecell[c]{Ours~(${\scriptstyle T=4}$)} & \makecell[c]{SS} & \makecell[c]{\textbf{86.7}} & \makecell[c]{85.9} & \makecell[c]{71.6} & \makecell[c]{68.6} & \makecell[c]{\textbf{78.3}} & \makecell[c]{72.5} & \makecell[c]{63.6} & \makecell[c]{75.3} \\
    \hline \toprule  
    \end{tabular}
    }
    \label{tab:posetrack2018-pose-evaluation1}
\end{table}

\begin{table}[ptb]
    \centering
    \caption{Quantitative evaluations (MOTA) of tracking on Posetrack2018 val set.}
    \setlength{\tabcolsep}{0.8mm}
    \resizebox{\columnwidth}{!}{
    \begin{tabular}{@{}c|c|cccccccc@{}}
    \bottomrule \hline
        \multicolumn{2}{c|}{Method} & \makecell[c]{Head} & \makecell[c]{Sho} & \makecell[c]{Elb} & \makecell[c]{Wri} & \makecell[c]{Hip} & \makecell[c]{Kne} & \makecell[c]{Ank} & \makecell[c]{Avg}\\
    \hline
        \makecell[c]{DetTrack(2020)~\cite{wang2020combining}} & \makecell[c]{TD} & \makecell[c]{74.2} & \makecell[c]{76.4} & \makecell[c]{\textbf{71.2}} & \makecell[c]{\textbf{64.1}} & \makecell[c]{64.5} & \makecell[c]{\textbf{65.8}} & \makecell[c]{\textbf{61.9}} & \makecell[c]{\textbf{68.7}} \\
        \makecell[c]{PT\_CPN++(2018)~\cite{yu2018multi}} & \makecell[c]{TD} & \makecell[c]{68.8} & \makecell[c]{73.5} & \makecell[c]{65.6} & \makecell[c]{61.2} & \makecell[c]{54.9} & \makecell[c]{64.6} & \makecell[c]{56.7} & \makecell[c]{64.0} \\
        \makecell[c]{Rajasegaran et al.(2021)~\cite{rajasegaran2021tracking}} & \makecell[c]{TD} & \makecell[c]{-} & \makecell[c]{-} & \makecell[c]{-} & \makecell[c]{-} & \makecell[c]{-} & \makecell[c]{-} & \makecell[c]{-} & \makecell[c]{55.8} \\
        \makecell[c]{Rajasegaran et al.(2022)~\cite{rajasegaran2022tracking}} & \makecell[c]{TD} & \makecell[c]{-} & \makecell[c]{-} & \makecell[c]{-} & \makecell[c]{-} & \makecell[c]{-} & \makecell[c]{-} & \makecell[c]{-} & \makecell[c]{58.9} \\
    \hline
        \makecell[c]{TML++(2019)~\cite{hwang2019pose}} & \makecell[c]{BU} & \makecell[c]{76.0} & \makecell[c]{76.9} & \makecell[c]{66.1} & \makecell[c]{56.4} & \makecell[c]{65.1} & \makecell[c]{61.6} & \makecell[c]{52.4} & \makecell[c]{65.7} \\
        \makecell[c]{STAF(2019)~\cite{raaj2019efficient}} & \makecell[c]{BU} & \makecell[c]{-} & \makecell[c]{-} & \makecell[c]{-} & \makecell[c]{-} & \makecell[c]{-} & \makecell[c]{-} & \makecell[c]{-} & \makecell[c]{60.9} \\
    \hline
        \makecell[c]{Ours~(${\scriptstyle T=1}$)} & \makecell[c]{SS} & \makecell[c]{82.0} & \makecell[c]{82.0} & \makecell[c]{58.8} & \makecell[c]{53.8} & \makecell[c]{72.3} & \makecell[c]{61.1} & \makecell[c]{40.2} & \makecell[c]{64.2} \\
        \makecell[c]{Ours~(${\scriptstyle T=4}$)} & \makecell[c]{SS} & \makecell[c]{\textbf{82.1}} & \makecell[c]{\textbf{82.3}} & \makecell[c]{59.0} & \makecell[c]{53.7} & \makecell[c]{\textbf{72.7}} & \makecell[c]{61.7} & \makecell[c]{41.7} & \makecell[c]{64.7} \\
    \hline \toprule  
    \end{tabular}
    }
    \label{tab:posetrack2018-tracking-evaluation}
\end{table}

We use Posetrack2018~\cite{PoseTrack} to validate that our method is flexible to 2D pose tracking task by simply skipping joint depth prediction. Since the provided annotations of PoseTrack2018 dataset is in 7.5 FPS, we downsample the input video to 7.5 FPS accordingly in both training and testing. We report our results on the validation set following prior works~\cite{wang2020combining,hwang2019pose,raaj2019efficient}. Tab.~\ref{tab:posetrack2018-pose-evaluation1} and~\ref{tab:posetrack2018-tracking-evaluation} present the results of Snipper on 2D pose estimation (AP) and pose tracking (MOTA). When comparing with bottom-up approaches, our method exceeds STAF~\cite{raaj2019efficient} for about 4\% in AP and MOTA, and also shows competitive results with TML++~\cite{hwang2019pose}. For the most recent top-down methods~\cite{wang2020combining}, our method is around 6\% AP and 4\% MOTA worse, which could attribute to the fact that our method jointly performs multi-person detection and pose tracking, while \cite{wang2020combining} relies on a specialized person detector to obtain the multi-person detection. When comparing with the most recent works~\cite{rajasegaran2021tracking,rajasegaran2022tracking}, our method still achieves around 8.9\% and 5.8\% MOTA better, which is largely credited to the inclusion of pose estimation that helps robust tracking.

\begin{figure*}[ptb]
    \centering
    \includegraphics[width=\textwidth]{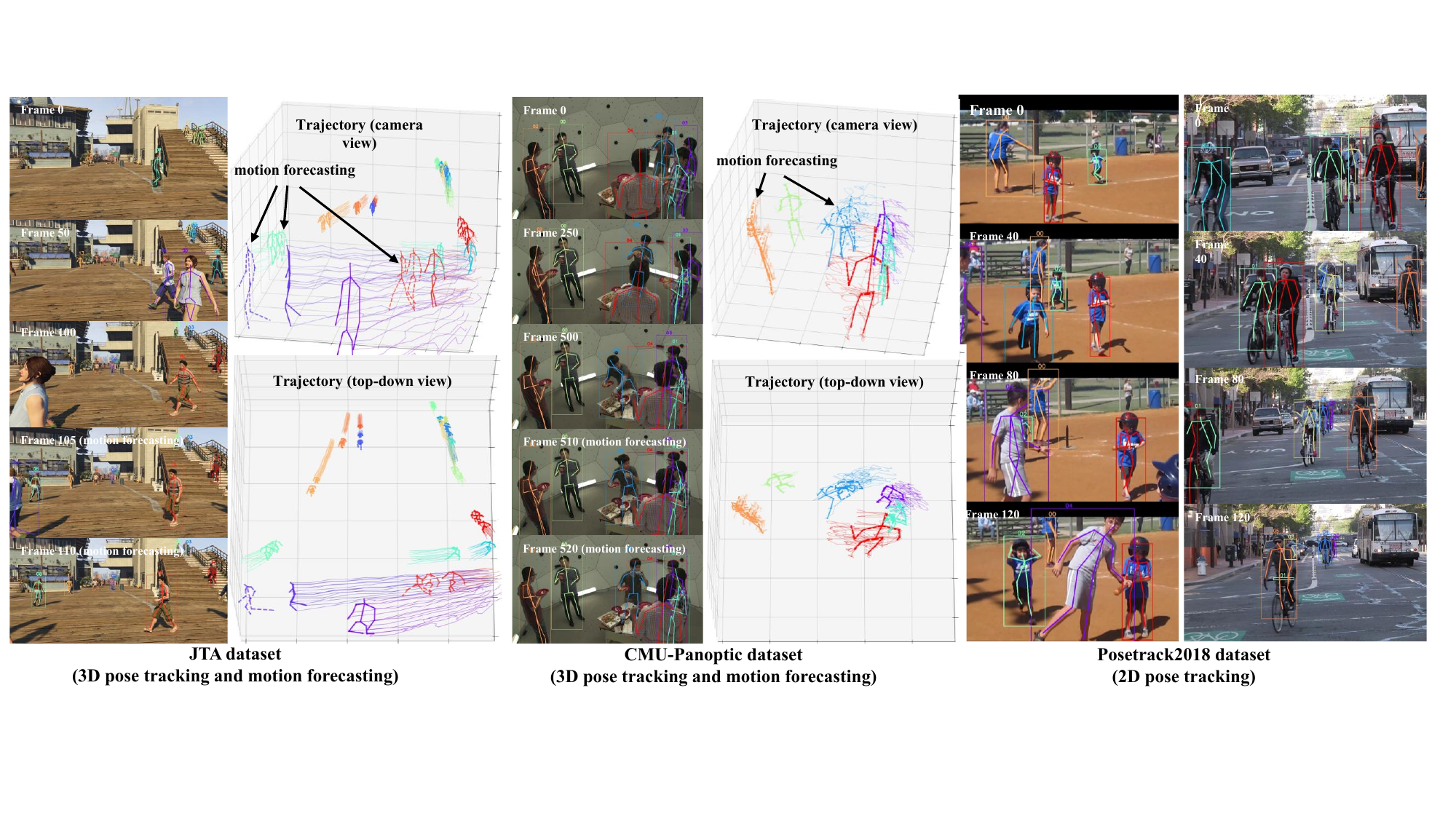}
    \caption{\textbf{Qualitative results on JTA, CMU Panoptic and Posetrack2018 datasets.} We show 3D pose tracking and motion forecasting results on JTA and CMU Panoptic datasets, and 2D pose tracking results on Posetrack dataset. Dashed 3D poses are the forecasting motion based on the last observed snippet.}
    \label{fig:vis-jta-panoptic}
\end{figure*}

\subsection{Ablation Study}
\label{sec:ablation}
\begin{figure}
    \centering
    \includegraphics[width=0.9\columnwidth]{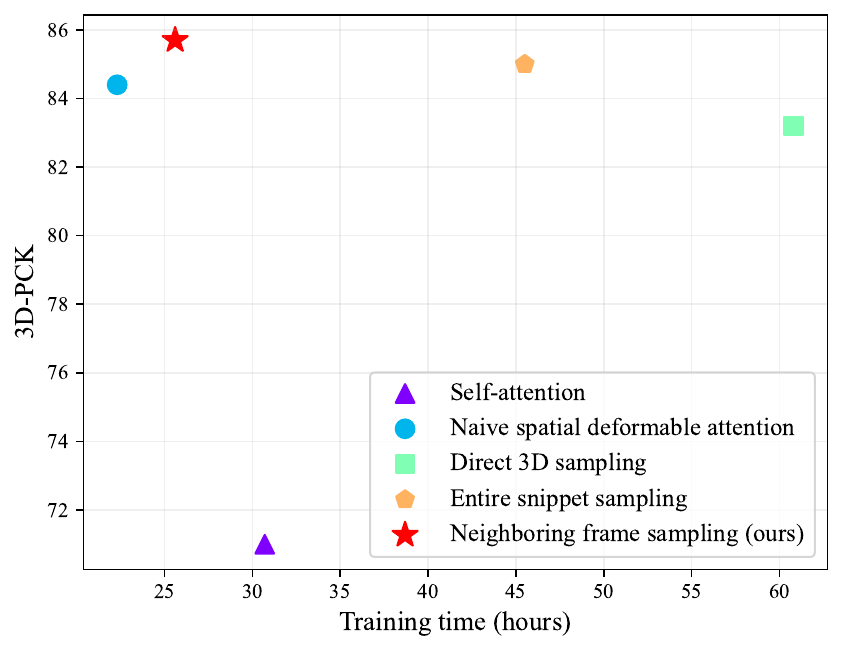}
    \caption{\textbf{Comparison of five attention strategies in terms of training time v.s. performance (3D-PCK).} We show the effectiveness and efficiency of our proposed spatiotemporal attention module. Training setting is identical for all five cases.}
    \label{fig:compare_attention}
\end{figure}

We present several key ablation studies (discussed in Sec.~\ref{sec:3Dattention}) on JTA dataset to highlight effectiveness and efficiency of our proposed spatiotemporal deformable attention and the correlation between pose estimation and tracking in Tab.~\ref{tab:ablation}. The comparison of five attention strategies in terms of training time v.s. performance is shown in Fig.~\ref{fig:compare_attention} to illustrate the effectiveness and efficiency of our proposed attention module. More ablation studies are provided in the supplementary.

\begin{table}[ptb]
    \centering
    \caption{Quantitative results of ablation study on JTA dataset.}
    \setlength{\tabcolsep}{1mm}
    \resizebox{\columnwidth}{!}{
    \begin{tabular}{@{}c|ccccc|c@{}}
    \bottomrule \hline
        & \multicolumn{5}{c|}{3D Pose Estimation} & \makecell[c]{Tracking} \\
    \cline {2-6}
        \multirow{-2}{*}{Method} & \makecell[c]{\footnotesize  AP} & \makecell[c]{\footnotesize  F1$@0.4$m} & \makecell[c]{\footnotesize  F1$@0.8$m} & \makecell[c]{\footnotesize  F1$@1.2$m} & \makecell[c]{\footnotesize  3D-PCK} & \makecell[c]{\footnotesize  MOTA} \\
    \hline
        \makecell[c]{Self attention} & \makecell[c]{53.2} & \makecell[c]{42.0} & \makecell[c]{55.0} & \makecell[c]{62.4}  & \makecell[c]{71.0} & \makecell[c]{49.9}\\
        \makecell[c]{Spatial deform. att.} & \makecell[c]{69.0} & \makecell[c]{53.8} & \makecell[c]{69.3} & \makecell[c]{75.3}  & \makecell[c]{84.4} & \makecell[c]{62.3}\\
        \makecell[c]{Direct 3D sampl.} & \makecell[c]{62.8} & \makecell[c]{54.3} & \makecell[c]{65.0} & \makecell[c]{71.6}  & \makecell[c]{83.2} & \makecell[c]{55.9}\\
        \makecell[c]{Entire snippet} & \makecell[c]{71.2} & \makecell[c]{59.3} & \makecell[c]{70.3} & \makecell[c]{76.2} & \makecell[c]{85.0}  & \makecell[c]{63.4}\\
    \hline
        \makecell[c]{Ours~(single scale)} & \makecell[c]{54.5} & \makecell[c]{43.1} & \makecell[c]{56.8} & \makecell[c]{64.1} & \makecell[c]{73.4} & \makecell[c]{50.2} \\
        \makecell[c]{Ours~(2D bbx tracking)} & \makecell[c]{66.5} & \makecell[c]{58.1} & \makecell[c]{69.0} & \makecell[c]{71.8} & \makecell[c]{81.9} & \makecell[c]{54.6} \\
        \makecell[c]{Ours~(2D pose tracking)} & \makecell[c]{67.1} & \makecell[c]{59.5} & \makecell[c]{69.3} & \makecell[c]{73.4}  & \makecell[c]{83.3} & \makecell[c]{55.9} \\
        \makecell[c]{Ours~(w/o smooth loss)} & \makecell[c]{69.1} & \makecell[c]{59.7} & \makecell[c]{70.7} & \makecell[c]{75.9}  & \makecell[c]{86.1} &  \makecell[c]{62.4}\\
    \hline
        \makecell[c]{Ours (1 FPS)} & \makecell[c]{68.5} & \makecell[c]{58.8} & \makecell[c]{69.1} & \makecell[c]{73.4}  & \makecell[c]{84.0} &  \makecell[c]{62.1}\\
        \makecell[c]{Ours (30 FPS)} & \makecell[c]{69.2} & \makecell[c]{59.2} & \makecell[c]{70.5} & \makecell[c]{75.2}  & \makecell[c]{84.2} &  \makecell[c]{62.7} \\
        \makecell[c]{Ours} & \makecell[c]{70.5} & \makecell[c]{60.3} & \makecell[c]{71.5} & \makecell[c]{76.4}  & \makecell[c]{85.7} &  \makecell[c]{63.2}\\
    \hline \toprule  
    \end{tabular}
    }
    \label{tab:ablation}
\end{table}

\boldstart{Self-attention}~\cite{wang2021end}. Our method outperforms self-attention (more than 20\% in all metrics), which is due to two main factors: (1) destruction of spatiotemporal local context within multi-frame features and (2) prohibitive compute for high-resolution multi-scale image features needed to estimate pose of small people. 

\boldstart{Na\"ive spatial deformable attention}~\cite{zhu2020deformable}.
This strategy performs slightly worse than ours, especially for accurate pose estimation (53.8 vs 60.3 of F1$@0.4$m) as it neglects scene/camera/object motions through time, where the temporal features at the same image position are not corresponding.

\boldstart{Direct 3D sampling.} 
This strategy also outperforms self-attention, showing that deformable attention in 3D space is an essential technique to encode spatiotemporal information from multi-frame features. However, it is still worse than ours by a large margin, mainly because the interpolation in temporal domain is ill-defined without known correspondences between frames such as optical flow, which leads to the aggregated features incorrect in time. 

\boldstart{Entire snippet sampling.} 
Our mechanism requires less computation cost and takes only 50\% training time of model with attention over the entire snippet, but gives similar performance, which showcases the efficiency of our approach.

\boldstart{Ours (2D pose tracking) and Ours (bbx tracking)} means the association between snippets is based on 2D pose or 2D bounding box instead of 3D pose, which reduces 7.3\% on MOTA and 2\% on 3D-PCK for 2D pose and 8.6\% and 3.4\% for 2D bounding box. Compared to 2D pose, 3D pose alleviates the issues of occlusion between people on an image since depth information is considered, which demonstrates that effective pose estimation facilitates tracking. 

\boldstart{Ours (w/o smooth loss)} denotes the model trained without joint smoothness loss $\mathcal{L}_{\text{smooth}}$ in Eq.~\ref{eq:hungarian-loss} for a snippet of 4 frames as the input. The pose scores are lower than our full model, illustrating that contextual information for tracking also helps improve pose estimation. However, this model without the smoothness loss still performs better than our model but with 1-frame input, indicating the effectiveness of the spatiotemporal attention model.

\boldstart{Ours (1 FPS) and Ours (30 FPS)} denote the model trained with video snippets at 1 FPS and 30 FPS respectively. The pose tracking performances are slightly worse than using 6 FPS, which can be the factor that very high frame rate introduces too much redundancy and easily results in overfitting during training while very low frame rate introduces too much discrepancy and results in missing clues for tracking and forecasting.

\begin{figure*}
    \centering
    \includegraphics[width=\textwidth]{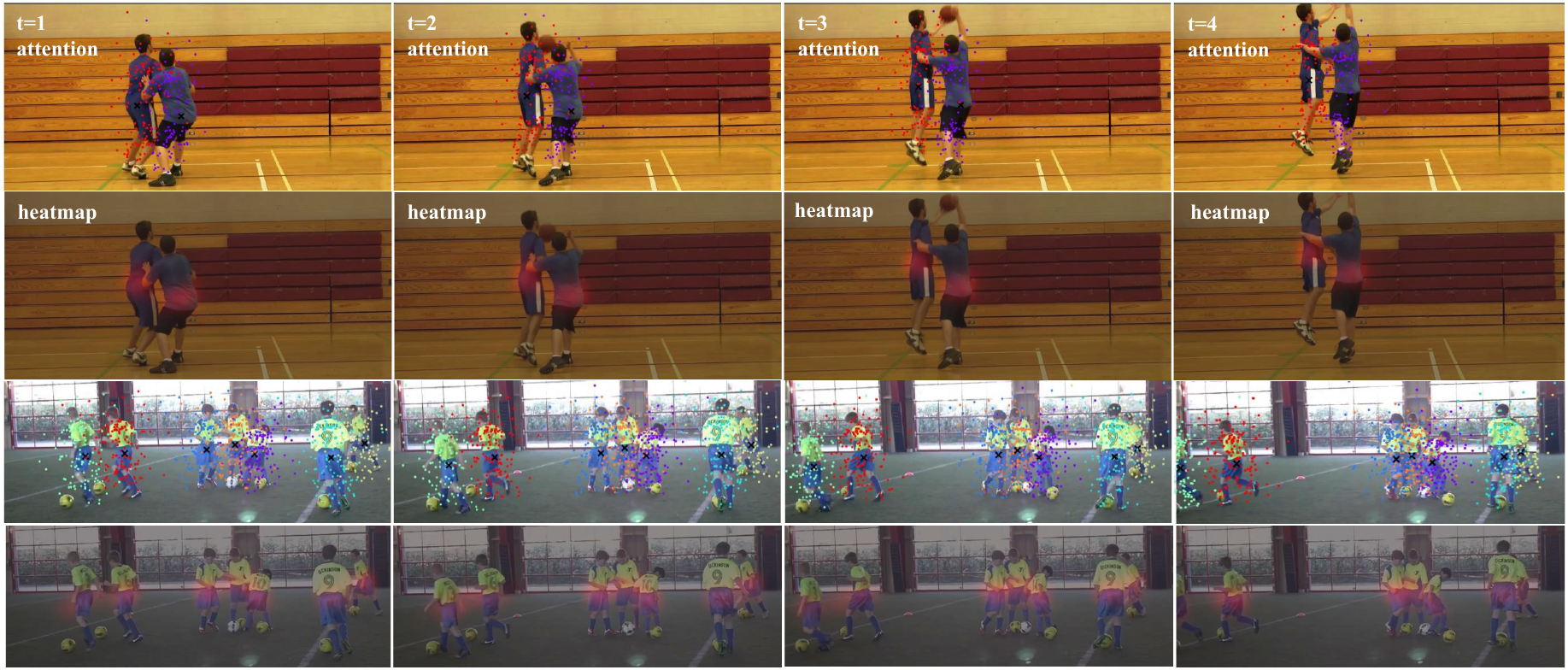}
    \caption{\textbf{Visualization of deformable attention in the transformer decoder  and heatmaps of root joint.} (best viewed in color) The sampling positions of the deformable attention for each person's pose query are displayed in the same color and ``$\times$" means the predicted root joint position. We show that the sampling positions of deformable attention usually cover the whole body of each person for effective and efficient pose feature accumulation.}
    \label{fig:vis_attention}
\end{figure*}

\begin{figure}
    \centering
    \includegraphics[width=\columnwidth]{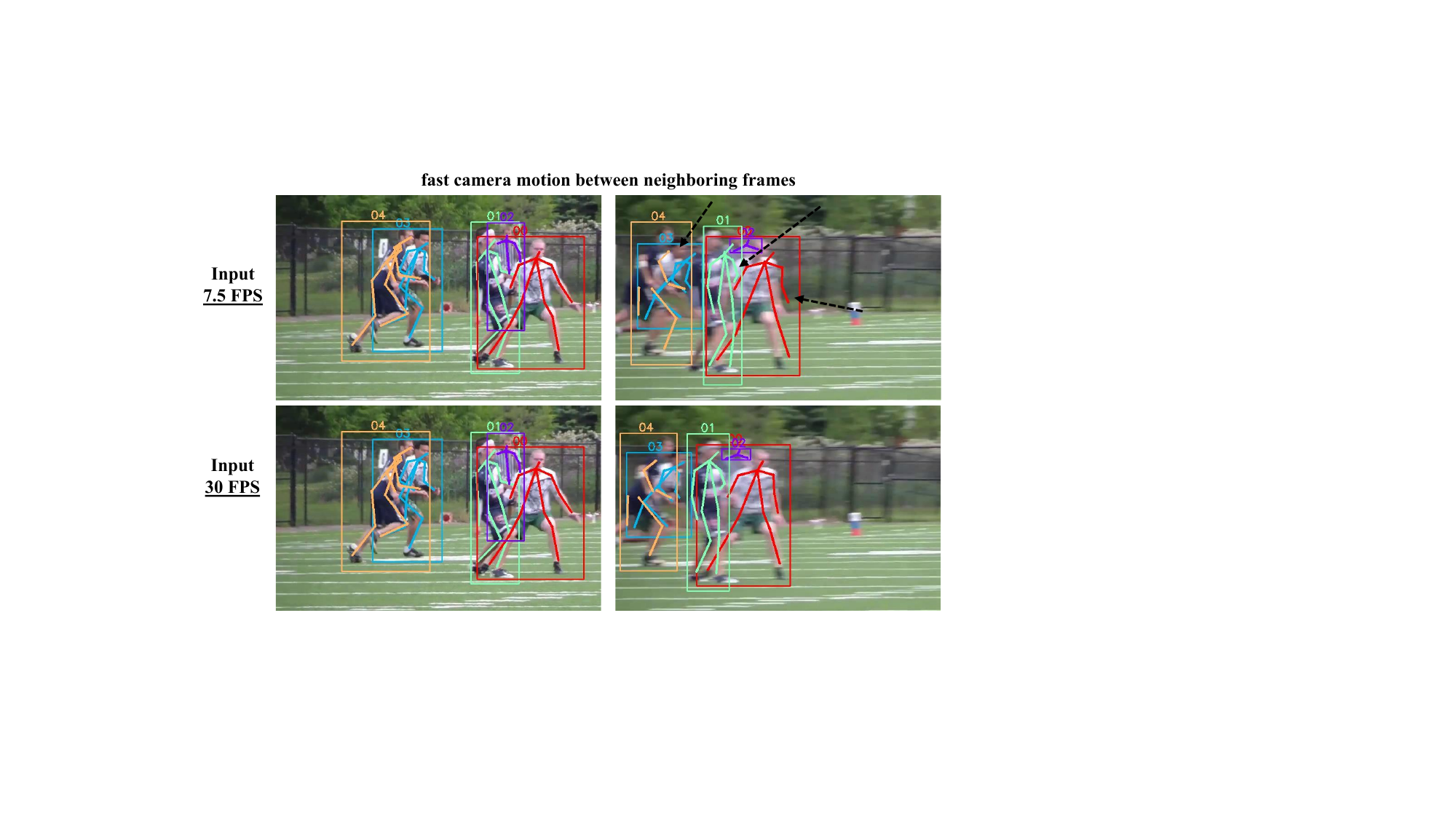}
    \caption{\textbf{Failure case.} Our method using video snippet at 7.5 FPS as input for inference could suffer from fast camera motion due to the large discrepancies between neighboring frames. In this special case, when we use video snippet at 30 FPS for inference, the model gives much better results due to the preserved correlation between frames in the input video, despite of fast camera motions.}
    \label{fig:failure}
\end{figure}

\subsection{Discussion}
\boldstart{Correlations among three tasks.}
Pose estimation and tracking are correlated with each other. The accurate 3D poses facilitate tracking, whereas robust tracking provides the informative temporal clues for pose estimation within the snippet. This is validated by comparing the consistent better pose estimation and tracking performance of Ours~(${\scriptstyle T=4}$) than Ours~(${\scriptstyle T=1}$) or multi-stage methods~\cite{fabbri2020compressed,tu2020voxelpose,wang2020hmor,rajasegaran2022tracking} on the three datasets in Tab.~\ref{tab:jta-evaluation},~\ref{tab:panoptic-evaluation1},~\ref{tab:panoptic-evaluation2},~\ref{tab:posetrack2018-pose-evaluation1} and~\ref{tab:posetrack2018-tracking-evaluation}.

On the other hand, pose tracking builds the crucial history for motion forecasting demonstrated by the better forecast motion of \textit{Ours} than no forecasting in Tab.~\ref{tab:JTA_future_motion}. Though in this paper, we cannot demonstrate that motion forecasting in turn helps pose tracking, We include the important task of motion forecasting in our unified framework for another two purposes: efficiency and robustness. For efficiency, the encoded spatiotemporal features within the video snippet capture crucial history for motion forecasting. Thus we can address the three tasks in a single stage with our unified framework for efficiency, i.e. \textit{running one network vs. many networks}. TesseTrack~\cite{reddy2021tessetrack} ($\sim$100M params) addresses only pose tracking, while our method (only 43M params) addresses all three tasks. For robustness, existing work on motion forecasting uses off-the-shelf pose estimators, such as HMP~\cite{cao2020long}, for the history pose estimation. But the pose estimators could fail unpredictably and thus cause forecasting to fail. Solving them jointly can be robust to pose tracking errors. We validate that our unified framework helps motion forecasting in Tab.~\ref{tab:JTA_future_motion}, where our method shows better performance than the multi-stage method HMP~(Hourglass~\cite{newell2016stacked}). 

\boldstart{Generalization ability.} To validate the generalization ability of our method, we train our model on a hybrid of datasets, MuCo~\cite{mehta2018single}, COCO~\cite{lin2014microsoft}, Posetrack2018~\cite{PoseTrack} and JTA~\cite{fabbri2018learning}. Then we test our model to predict 3D pose tracking on the unseen MuPoTS~\cite{mehta2018single} and Posetrack val set (no 3D pose annotations). The qualitative results are included in the supplementary video, where the predicted 3D pose is smooth and the tracking is consistent across the entire videos, even in occlusion cases.

\boldstart{Attention visualization.} We visualize the attention maps of the last layer in the transformer decoder in Fig.~\ref{fig:vis_attention}, where the sampling positions for each person's pose queries are presented by the same color and ``$\times$" means the root joint position. Our proposed spatiotemporal deformable attention samples around each person's whole body and later aggregates these sampled features together to update pose queries. Compared with self-attention, our proposed deformable attention typically preserves better spatial and temporal relationships for correct pose regression, and compared with the na\"ive spatial deformable attention, our method considers the motion changes between frames as is indicated by the root joint positions "$\times$" of each person in the video snippet in Fig.~\ref{fig:vis_attention}.

\boldstart{Failure case and limitations.} Our method could suffer frame fast camera motion as shown in Fig.~\ref{fig:failure}, where there are quite large discrepancies between neighboring frames at 7.5 FPS. In this special case, the problem can be alleviated when we use video snippet at 30 FPS as the input due to the preserved correlation between frames. Another limitation lies in the spatiotemporal positional encoding that does not consider the correspondence between frames, especially the camera motion exists between frames. Future work could explore explicitly use the camera poses such as ray-based position encoding and the low-resolution optical flow to guide the feature aggregation.

\section{Conclusion}
We present Snipper, a unified spatiotemporal Transformer for simultaneous multi-person 3D pose estimation, tracking and motion forecasting on a video snippet. We propose an efficient yet powerful spatiotemporal deformable attention module to aggregate spatiotemporal information across the snippet. We demonstrate the effectiveness of Snipper on three challenging public datasets where a generic Snipper model rivals specialized state-of-art baselines trained for 3D pose estimation, tracking, and forecasting.


%


\section*{Acknowledgment}
Thank Wei Ji and Jingjing Li for their constructive advice. This work was partially supported during Shihao Zou's internship at Meta Reality Labs. The subsequent portion of this work was funded by the University of Alberta Start-up Grant and the NSERC Discovery Grant No. RGPIN-2019-04575. Shihao Zou was also supported by Alberta Innovates Graduate Student Scholarship for Data-Enabled Innovation.

\ifCLASSOPTIONcaptionsoff
  \newpage
\fi



\bibliographystyle{IEEEtran}
\bibliography{IEEEabrv}
\end{document}


%
\title{Supplementary Material\\Snipper: A Spatiotemporal Transformer for Simultaneous Multi-Person 3D Pose Estimation Tracking and Forecasting on a Video Snippet}
%
%
%

\author{Shihao~Zou,
        Yuanlu~Xu,
        Chao~Li,
        Lingni~Ma, 
        Li~Cheng,
        and~Minh~Vo
\thanks{Part of this work was done during Shihao Zou's internship at Meta Reality Labs. The results on JTA dataset were added after his internship. Minh Vo was at Meta Reality Labs during the development of this work.}
\thanks{S. Zou and L. Cheng are with the Department
of Electrical and Computer Engineering, University of Alberta, Canada. (E-mail: szou2@ualberta.ca, lcheng5@ualberta.ca).}
\thanks{Y. Xu, C. Li, and L. Ma are with the Meta Reality Labs, United States. (E-mail: yuanluxu@meta.com, chao.li@meta.com, lingni.ma@meta.com).}
\thanks{M. Vo is with Spree3D, United States. (E-mail: minh.vo@spree3d.com).}
\thanks{Manuscript received 08 Oct 2022; accepted 03 Feb 2023. Li Cheng is the corresponding author for this paper.}}

%
%

\markboth{Journal of \LaTeX\ Class Files,~Vol.~14, No.~8, August~2015}%
{Shell \MakeLowercase{\textit{et al.}}: Bare Demo of IEEEtran.cls for IEEE Journals}
%



\maketitle

%
\IEEEpeerreviewmaketitle

\section{Encoder with Multi-layer Attention}
\begin{figure}[ptb]
    \centering
    \includegraphics[width=0.75\columnwidth]{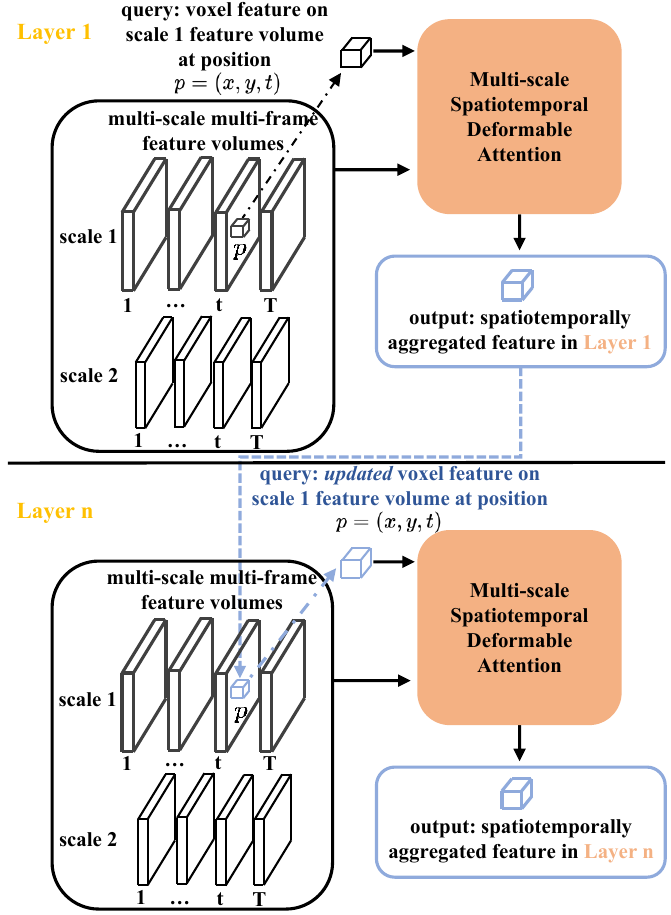}
    \caption{\textbf{Multi-layer Transformer Encoder.} In the $1$-st layer of attention, consider the voxel at position (x, y, t) of scale 1 feature volume as an example, we use this voxel feature as the query in the attention module to aggregate features from multi-scale feature volumes, and obtain the spatiotemporally aggregated feature to replace the old voxel feature at position (x, y, t) of scale 1 feature volume. The same process is executed for \emph{all the voxels in all-scale feature volumes}. The updated multi-scale feature volumes are then fed to the next layer of attention module iteratively. 
    }
    \label{fig:multilayer-encoder}
\end{figure}

Our transformer encoder stacks multiple layers of attention module. The input is multi-scale multi-frame image features extracted by CNN. In each layer, \emph{all the voxels in all-scale feature volumes} act as the query to aggregate multi-scale spatiotemporal features. The updated feature volumes are used as the input to the next layer of attention module iteratively. The process is summarized in Fig.~\ref{fig:multilayer-encoder}.

\section{Decoder with Multi-layer Attention}
\begin{figure}[ptb]
    \centering
    \includegraphics[width=0.9\columnwidth]{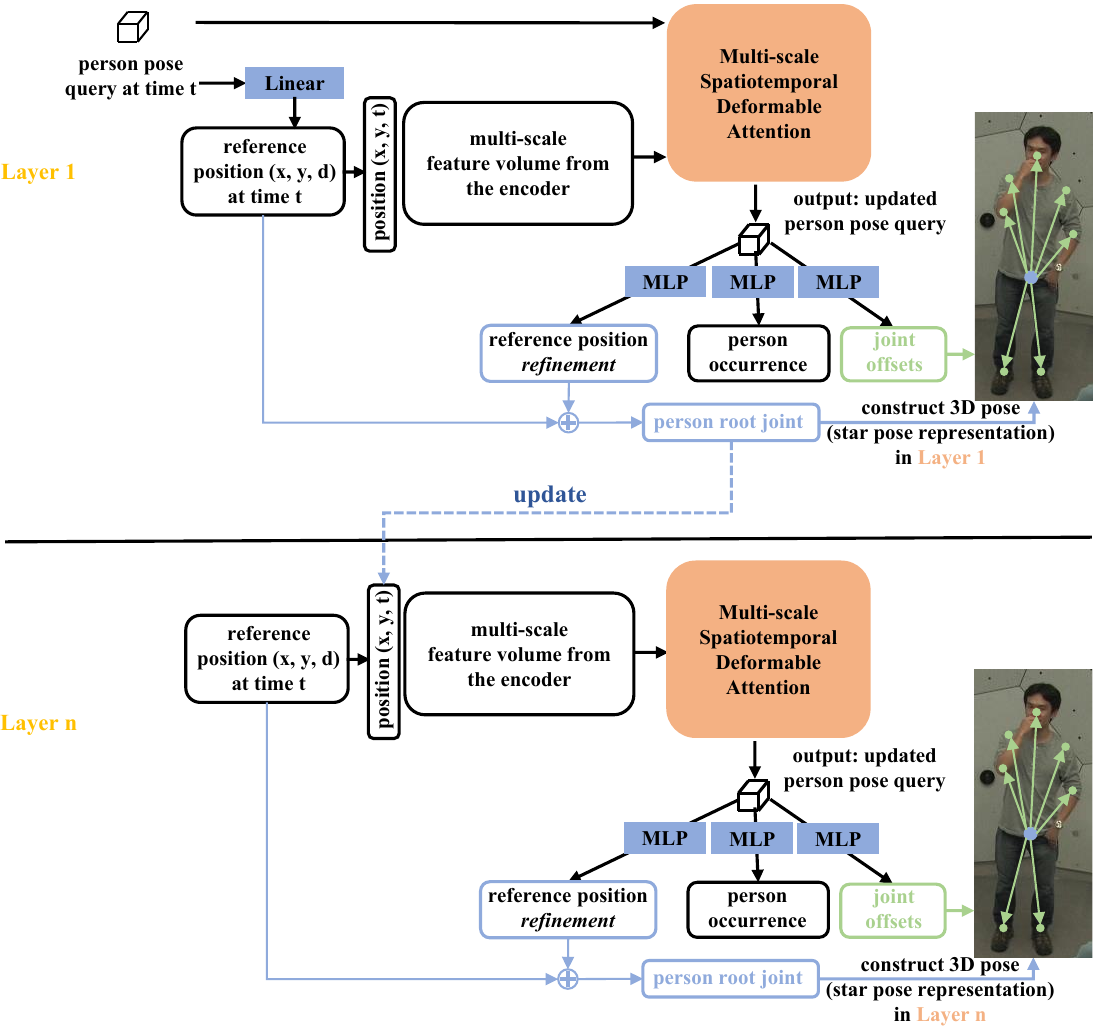}
    \caption{\textbf{Multi-layer Transformer Decoder.} In the $1$-st layer, a person pose query $q_t^0$ specified at time $t$ regresses a reference position $(x^{1}, y^{1}, d^{1})$ to form the sampling position $(x^{1}, y^{1}, t)$ in the multi-scale feature volumes. Then it acts as the query in the attention module to obtain the updated person pose query $q_t^1$ in the $1$-st layer. This updated query is used to regress to the joint offsets $\mathbf{V}^1$, probability of person occurrence $o^1$ and position refinement $(\Delta x^1, \Delta y^1, \Delta d^1)$. 
    On the one hand, the reference position $(x, y, d)$ is regarded as the person root joint position, and together with the joint offsets, the 3D pose in the $1$-st layer $\mathbf{P}_t^1$ can be constructed.
    On the other hand, the position refinement is used to update the reference position in the next layer of attention, $(x^2, y^2, d^2)=(x^{1} + \Delta x^{1}, y^{1} + \Delta y^{1}, d^{1} + \Delta d^{1})$.
    For the $n$-th layer, the query to the attention module becomes the updated person pose query from the last layer $q_t^{n-1}$, and the reference position is updated via $(x^n, y^n, d^n)=(x^{n-1} + \Delta x^{n-1}, y^{n-1} + \Delta y^{n-1}, d^{n-1} + \Delta d^{n-1})$ and form the new sampling position $(x^{n}, y^{n}, t)$. After obtaining the updated pose query, the 3D pose in the $n$-th layer $\mathbf{P}_t^n$ can be constructed.
    }
    \label{fig:multilayer-decoder}
\end{figure}

Our transformer decoder stacks multiple layers of attention module. The input is $N(T+T_f)$ person pose queries, which are used to regress $N$ people's 3D pose trajectories for $T$ observed frames and $T_f$ future frames. In each attention layer, these pose queries accumulate pose features and reconstruct 3D poses iteratively.
In the $1$-st layer, assuming a person pose query at time $t$ is $q_t^0$, we use it to regress a reference position $(x^{1}, y^{1}, d^{1})$ and feed it into the attention module as the query to aggregate pose features at the sampling position $(x^{1}, y^{1}, t)$ in multi-scale feature volumes. The output is the updated person pose query $q_t^1$ for the $1$-st layer, which regresses the joint offsets and occurrence probability to construct 3D pose in the $1$-st layer $\mathbf{P}_t^1$, as well as position refinement $(\Delta x^1, \Delta y^1, \Delta d^1,)$ to update the reference position in the next layer. For the $n$-th layer, the attention module takes the updated person pose query from the last layer $q_t^{n-1}$, and aggregates pose features at the position $(x^n, y,^n d^n)=(x^{n-1} + \Delta x^{n-1}, y^{n-1} + \Delta y^{n-1}, d^{n-1} + \Delta d^{n-1})$. Similarly, the updated pose query for the $n$-st layer is regressed to construct the 3D pose $\mathbf{P}_t^n$ in the $n$-th layer. The process is summarized in Fig.~\ref{fig:multilayer-decoder}.

\section{Depth and Joint Offset Normalization}
We assume the camera intrinsic is $(f_c, c_x, c_y)$ where $f_c$ is the focal length and $(c_x, c_y)$ is the center of image. According to the pinhole camera model, we have $x = \frac{X}{d} \cdot f_c + c_x$ and $y =\frac{Y}{d} \cdot f_c + c_y$, where $(X, Y, d)$ is the 3D position and $(x, y)$ is the projected 2D position on the image. We can avoid predicting the focal length by normalizing the depth $d$ with $f_c$, \ie $\tilde d = d / f_c$.

The joint offset $(\Delta x, \Delta y)$ is represented in the image pixel distance, which will become smaller if the person moves far away from the camera. According to the pinhole camera model, we have $\Delta x = \frac{\Delta X}{d} \cdot f_c$ and $\Delta y = \frac{\Delta Y}{d} \cdot f_c$, which shows that the magnitude of the 2D offset in pixel distance is proportional to $f_c/d$. Therefore, we propose to normalize the joint offsets with the normalized depth, \ie $\Delta \tilde x = \Delta x \cdot \tilde d$ and $\Delta \tilde y = \Delta y \cdot \tilde d$. Then, $(\Delta \tilde x, \Delta \tilde y)$ has the identical magnitude to the offset $(\Delta X, \Delta Y)$ in 3D space. Thus, the magnitude of 2D normalized joint offset only depends on the pose of the person, which is more consistent across identities.

\section{Ablation Study}
We present more ablation studies on JTA dataset to highlight the effectiveness of some key components in our method in Tab.~\ref{tab:ablation}.

\begin{table}[]
    \centering
    \caption{Quantitative results of ablation study on JTA dataset.}
    \setlength{\tabcolsep}{0mm}
    \resizebox{0.5\textwidth}{!}{
    \begin{tabular}{c|p{35pt}p{40pt}p{40pt}p{40pt}p{40pt}|p{40pt}}
    \bottomrule \hline
        & \multicolumn{5}{c|}{3D Pose Estimation} & \makecell[c]{Tracking} \\
    \cline {2-7}
        \multirow{-2}{*}{Method} & \makecell[c]{AP} & \makecell[c]{F1$@0.4$m} & \makecell[c]{F1$@0.8$m} & \makecell[c]{F1$@1.2$m} & \makecell[c]{3D-PCK} & \makecell[c]{MOTA} \\
    \hline
        \makecell[c]{No temp. enc.} & \makecell[c]{67.3} & \makecell[c]{56.5} & \makecell[c]{68.3} & \makecell[c]{73.7}  & \makecell[c]{84.9} & \makecell[c]{55.0}\\
        \makecell[c]{Trajectory query} & \makecell[c]{69.3} & \makecell[c]{58.6} & \makecell[c]{71.1} & \makecell[c]{76.3}  & \makecell[c]{85.0} &  \makecell[c]{62.9}\\
        \makecell[c]{W/o offset norm.} & \makecell[c]{69.1} & \makecell[c]{57.1} & \makecell[c]{70.8} & \makecell[c]{75.2}  & \makecell[c]{78.5} &  \makecell[c]{62.4}\\
        \makecell[c]{Encoder layers 4} & \makecell[c]{67.1} & \makecell[c]{58.0} & \makecell[c]{69.3} & \makecell[c]{74.4}  & \makecell[c]{85.3} &  \makecell[c]{60.6}\\
        \makecell[c]{Encoder layers 2} & \makecell[c]{61.8} & \makecell[c]{52.2} & \makecell[c]{63.6} & \makecell[c]{69.1} & \makecell[c]{81.7} &   \makecell[c]{53.4}\\
        \makecell[c]{W/o heatmap} & \makecell[c]{68.8} & \makecell[c]{57.4} & \makecell[c]{70.1} & \makecell[c]{74.8} & \makecell[c]{82.6} &   \makecell[c]{61.6}\\
    \hline
        \makecell[c]{Ours} & \makecell[c]{70.5} & \makecell[c]{60.3} & \makecell[c]{71.5} & \makecell[c]{76.4}  & \makecell[c]{85.7} &  \makecell[c]{63.2}\\
    \hline \toprule  
    \end{tabular}
    }
    \label{tab:ablation}
\end{table}

\boldstart{No temporal encoding} is proposed to make the pose queries aware of the chronological order within a trajectory. Without temporal encoding, the pose tracking accuracy decreases by over 8\% on MOTA, which may attribute to more frequent person ID switching without adding trajectory temporal encoding. 

\boldstart{Trajectory query.} In Snipper, there are $N(T+T_f)$ queries with each query focusing on \emph{the pose of each queried person at each time}. To illustrate the effectiveness of $N(T+T_f)$ queries strategy, we compare it with the strategy of $N$ queries where each query focuses on \emph{the trajectory of each queried person}. We can see from Tab.~\ref{tab:ablation} that $N$ queries strategy produces worse accuracy of pose tracking (69.3 vs. 70.5 for AP and 62.9 vs. 63.2 for MOTA) since there is bottle-neck between the dimension of query embedding (each embedding is of size 384) and pose trajectory (each $T=4$ trajectory is of size 240 with 15 3D joints and visibility), especially for large $T$.

\boldstart{W/o offsets normalization.} For 3D pose estimation, we propose to normalize 2D joint offsets by the depth to overcome the issue of scale on 2D image. The accuracy reduction, especially 3D-PCK (-7.2\%), illustrates the effectiveness of our proposed normalization strategy.

\boldstart{Number of layers of encoder.} The transformer encoder is the key component to encode spatiotemporal features of the input snippet. The number of layers in the encoder is able to illustrate the effectiveness of spatiotemporal deformable attention to encode this feature, as is shown in Tab.~\ref{tab:ablation}.

\boldstart{Heatmap supervision} This corresponds to supervising the first $N_J$ channels of each temporal slice in the volume to be the multi-person joints heatmap, as described in ln-310 in the main text. As can be seen in Tab.~\ref{tab:ablation}, this intermediate supervision improves the 3D pose accuracy (2.9\% of 3D-PCK).
